\documentclass{bmvc2k}

\addauthor{Ahmad Darkhalil}{ahmad.darkhalil@bristol.ac.uk}{1}
\addauthor{Dima Damen}{dima.damen@bristol.ac.uk}{1}
\addauthor{David Fouhey}{david.fouhey@nyu.edu}{2}

\addinstitution{
 School of Computer Science\\
 University of Bristol, Bristol, UK
}
\addinstitution{
 Computer Science and\\
 Electrical and Computer Engineering\\
 New York University, NY, US
}

\runninghead{Darkhalil et al}{Improving .. Hand-Object Interaction Detection}

\usepackage{comment}
\usepackage{amssymb}
\usepackage{booktabs}
\usepackage{soul}
\usepackage{cleveref}
\usepackage{wrapfig}
\usepackage{multirow}
\usepackage{colortbl}
\usepackage{enumitem}

\usepackage{graphicx}
\usepackage{amsmath}
\usepackage{pifont}  
\usepackage{shortbold}

\usepackage{amssymb}
\usepackage{booktabs}
\usepackage{soul}
\usepackage{cleveref}
\usepackage{wrapfig}
\usepackage{multirow}
\usepackage{colortbl}
\usepackage{enumitem}
\usepackage{placeins}
\usepackage{adjustbox}
\usepackage{xcolor}

\usepackage{graphicx}
\usepackage{amsmath}
\usepackage{pifont}  
\usepackage{shortbold}  

\begin{document}
\newcommand{\method}{HOI-DETR\xspace}
\newcommand{\ToDo}[2]{{\noindent \textcolor{red}{\textbf{#1}}: \textcolor{red}{#2}}}
\definecolor{myred}{RGB}{220,50,32}
\definecolor{myyellow}{RGB}{200,150,0}
\definecolor{mycyan}{RGB}{0,159,115}
\definecolor{mygray}{RGB}{80,80,80}

\title{Improving and Evaluating Hand-Object Interaction Detection} 

\maketitle

\begin{abstract}
Understanding hands and the objects they interact with, both directly and through tools, is a key step for tasks ranging from action perception to 3D reconstruction and robotics. Our paper provides several contributions to the Hand-Object Interaction (HOI) understanding literature: (1) HOI-DETR, a new framework that introduces hand-object and object-object interactions to the Co-DETR architecture to produce a state-of-the-art method; (2) a comprehensive HOI evaluation suite of 4 diverse datasets, including a video benchmark derived from the HD-EPIC dataset and fresh annotations that improve the Hands23 benchmark and (3) a trained checkpoint that significantly improves the state of the art across Hands23, HOIST, FineBio, and HD-EPIC, including mAP gains of over 20 percentage points on Hands23 and FineBio. Our ablations confirm the contributions of each model component.
Qualitative results of HOI-DETR are shown in Figure~\ref{fig:intro}.
Code, model, and datasets are available on our \href{https://ahmaddarkhalil.github.io/HOI-DETR/}{project page:  https://ahmaddarkhalil.github.io/HOI-DETR/}.
\vspace*{-16pt}

\end{abstract}
  
\begin{figure}[t]
    \centering
    \includegraphics[width=\textwidth]{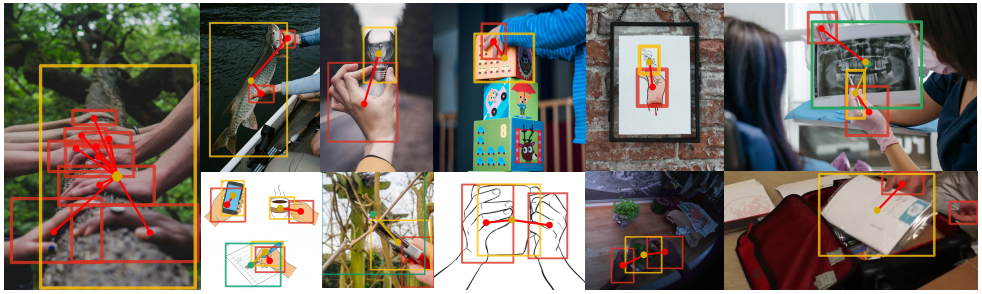}
    \vspace*{-16pt}
    \caption{We propose \textbf{\method}, a method that employs \textbf{DE}tection \textbf{TR}ansformers for the \textbf{H}and-\textbf{O}bject \textbf{I}nteraction task.
    HOI-DETR predicts all visible \textcolor{myred}{hands}, \textcolor{myyellow}{1st objects} (objects in direct physical interaction with a hand), and \textcolor{mycyan}{2nd objects} (objects that the \textcolor{myyellow}{1st objects} acts upon when it is used as a tool). \method also predicts interaction, i.e.\ \textcolor{myred}{hands}~$\to$~\textcolor{myyellow}{1st objects} and \textcolor{myyellow}{1st objects}~$\to$~\textcolor{mycyan}{2nd objects} mappings for any (and all) instances in the input image. We show diverse variety of zero-shot examples (we select 9 challenge web images (some out of domain like drawings) and a couple of images from HOIST~\cite{sn_hoist_cvpr_2024} and HD-EPIC~\cite{perrett2025hdepic}), showcasing different hand/object sizes and viewpoints.}
    \vspace*{-12pt}
    \label{fig:intro}
\end{figure}

\section{Introduction}
\label{sec:intro}

Hands play a central role in how humans interact with the world. Accordingly, recognising hands and the objects that they are interacting with (hand-object interaction, or HOI) is an important step for many vision analyses. For instance, many methods for 3D hand~\cite{pavlakos2024reconstructing,prakash2024mitigating} and object~\cite{cao2021reconstructing} reconstruction begin by first locating the hand and the held object. Similarly, hand contact is key to tracking object movements~\cite{darkhalil2022epic,Plizzari2025OSNOM}. A full understanding of hand-object interaction typically involves locating and understanding the relationship between several key categories: hands, objects that hands interact with directly, and objects that hands interact with using held objects (typically tools).

\begin{figure}[t]
    \centering
    \includegraphics[width=\linewidth]{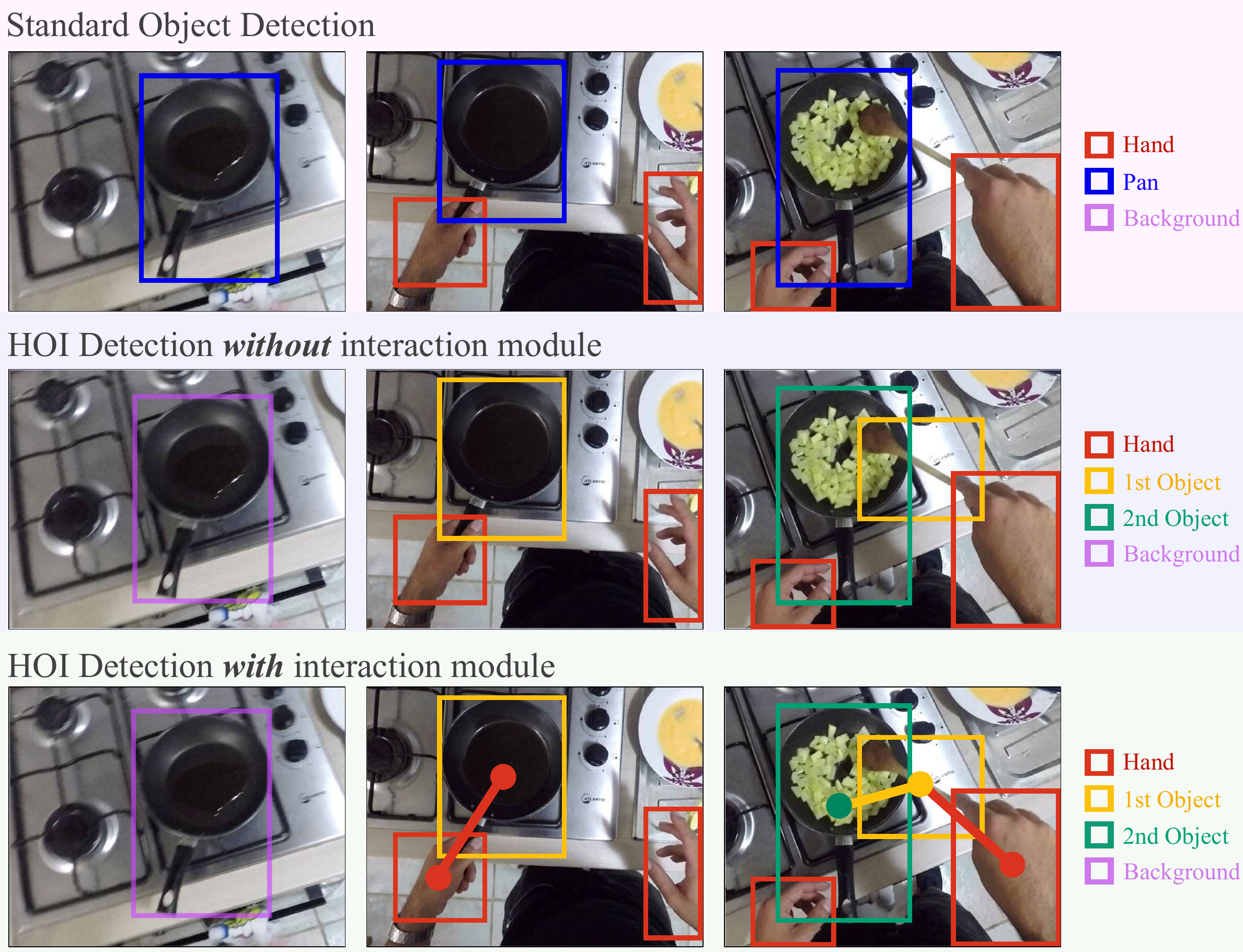}
    \vspace*{-16pt}
    \caption{Standard object detection vs HOI detection. \text{Top:} In standard detect tasks, object bounding boxes are detected based on their semantics.
    \text{Middle:} Unlike standard deduction, detections are dynamically defined by objects' interaction roles, the ``pan'' shifts from background (left), to a \textcolor{myyellow}{1st object} (middle), or a \textcolor{mycyan}{2nd object} (right). \text{Bottom:} In addition to HOI detections, explicit links connect the \textcolor{myred}{hands} to \textcolor{myyellow}{1st objects} and \textcolor{myyellow}{1st objects} to \textcolor{mycyan}{2nd objects.}}
    \vspace*{-12pt}
    \label{fig:back12example}
\end{figure}

This formulation highlights a critical distinction: the HOI task fundamentally reshapes the behaviour required from a standard object detector. As illustrated in Figure~\ref{fig:back12example}, standard detection focuses on object semantic classes (e.g., identifying a ``pan'', regardless of its role in any interactions). In contrast, HOI detection forces the model to understand contextual \emph{roles} with respect to the~\textcolor{myred}{hands} and other interacting objects. Depending on the ongoing interaction, the same pan should be classified as background when not involved in an interaction,  as a~\textcolor{myyellow}{1st object} if grasped/touched directly, or as a~\textcolor{mycyan}{2nd object} if manipulated via a tool. Consequently, a comprehensive HOI system must go beyond standard object localisation to dynamically assign state-driven roles and predict the interaction links.

Progress in this area, however, has been uneven. Some methods in hand detection~\cite{potamias2025wilor} have taken advantage of advances in generic object detection, but do not provide an understanding of the held objects. Others improve the tracking of held objects over time~\cite{sn_hoist_cvpr_2024} but do not identify which hand is holding the object. Methods that provide an understanding of the objects that hands are interacting with~\cite{shan2020understanding,cheng2023towards} have relied on older and less effective detectors~\cite{he2017mask}. In evaluation, while methods are often used on videos in practice~\cite{Damen2022RESCALING,pavlakos2024reconstructing,sn_hoist_cvpr_2024}, hand detection is usually solely evaluated on image-based metrics like mAP~\cite{shan2020understanding,narasimhaswamy2020detecting,cheng2023towards,potamias2025wilor} and do not consider spatiotemporal consistency. 

In our paper, we improve hand-object interaction detection in terms of methods, data, and evaluation. We introduce a new method, \underline{H}and-\underline{O}bject \underline{I}nteraction \underline{D}etection \underline{T}ransformer (HOI-DETR), that produces a full understanding of HOI with modern detection approaches. We compare HOI-DETR to the state-of-the-art with a new comprehensive benchmark, including video metrics to measure spatiotemporal consistency. Our contributions are:

\vspace{1mm}
\par \noindent {\bf Method.} We introduce HOI-DETR, which integrates an interaction module into the Co-DETR architecture~\cite{zong2023detrs} to enable modeling hand-object and object-object interactions. When trained end-to-end on Hands23~\cite{cheng2023towards}, HOI-DETR substantially improves the state of the art: mAP increases on two datasets by over 20 percentage points, reducing the state of the art~\cite{cheng2023towards} error rate by $52\%$ and outperforming a video-based method~\cite{sn_hoist_cvpr_2024} despite operating on frames.

\vspace{1mm}
\par \noindent {\bf Data.} We provide refined training data and a comprehensive evaluation suite combining annotations from four HOI datasets. These span exocentric and egocentric, images and videos, and include out-of-distribution data like FineBio~\cite{yagi2025finebio}. Additionally, we correct inconsistencies in the Hands23~\cite{cheng2023towards} dataset (training and evaluation), collecting new annotations for this dataset, as well as propose a new video-level benchmark (HD-EPIC-HOI) based on challenging clips from the newly released egocentric HD-EPIC~\cite{perrett2025hdepic} dataset.

\vspace{1mm}
\par \noindent {\bf Evaluation.} HOI detectors are commonly used on videos, but previous methods lack the necessary consistency. We do evaluation to produce more accurate assessments of the effectiveness of HOI detectors. We also ablate the individual subcomponents of HOI-DETR to identify what contributes to its superior performance.

\section{Related work}
\label{sec:related_work}

Our work introduces a new model and improved evaluation for hands engaged in interaction in the wild. It accordingly touches on several research areas.

\noindent\textbf{Hand Detection.}
Detecting hands is a long standing problem in computer vision. After early efforts (e.g., ~\cite{Jones2002StatisticalColor,mittal2011hand}), there were substantial advances with deep learning~\cite{bambach2015egohands}. These have to led approaches that tackle hands by themselves~\cite{potamias2025wilor}, often adapting recent advances in general detection. Hand detections are a key pre-processing step for subsequent analysis of the hands~\cite{simon2017hand,zimmermann2017learning,baek2018pushing,zimmermann2019freihand,mueller2018ganerated}. Our work follows this trend by providing a new, effective detection system, but our primary focus is detecting both hands {\it and} hand-held objects.

\noindent\textbf{Hand-Object Detection.} A critical extension of hand detection was identifying held objects and their relationship to hands. In these works, one trains a detector to recognise ``held object'' as a class, irrespective of the category: a pan that is held and a cup that is held both belong to the ``held object'' category, but are a non-detection when put down. While similar to human-object interaction~\cite{Sadhu_2021_CVPR, chao2018learning, gao2018ican, gkioxari2018detecting}, it differs by not naming the category of object (e.g., ``pan''). Early benchmarks, like 100 Days of Hands~\cite{shan2020understanding} provided detailed annotations, leading to new applications. This effort was scaled in Hands23~\cite{cheng2023towards}, which proposed both a detector and a dataset. 
We contribute the first transformer-based HOI detection model, superior in performance, equipped with a new interaction module and careful experiments that demonstrate the role of components.

\noindent\textbf{Hand-Object Segmentation, 3D Reconstruction, Keypoints.} There are many other related tasks that are complementary to our proposed methods and evaluations. These include hand-object segmentation~\cite{zhang2022fine, su2025care,su2026interaction}, or more specifically identifying the spatial extent of hands and held objects. This line of work is orthogonal and complementary to our efforts at improving and evaluating detection of hands, hand-held objects and their interaction. With the rise in many highly-effective systems (and foundation models) that generically convert boxes to segments~\cite{ravi2025sam,kirillov2023segment,ke2023segment}, we believe that the primary challenge is identifying the {\it held} object; once identified and boxed, it can be segmented easily. Similarly, there have been efforts for obtaining accurate keypoints~\cite{potamias2025wilor} and 3D meshes~\cite{pavlakos2024reconstructing,prakash2024mitigating}. These are also complementary since our proposed system provides better boxes for these downstream tasks.

\noindent\textbf{Video-Level Hand-Object Interaction Analysis.}
One of our contributions is spatiotemporal consistency measurements. This is part of a broader trend of moving towards analysing results over time. This was explored for hands by HOIST~\cite{sn_hoist_cvpr_2024}. Compared to HOIST, HOI-DETR additionally provides hand-object and object-object associations.

\noindent\textbf{Applications of Hand-Object Interaction.}
Understanding 2D hand-object interaction serves as a critical prior for many downstream tasks, e.g., for 3D reconstruction of hands~\cite{potamias2025wilor, pavlakos2024reconstructing, prakash2024mitigating}, grasping from observation~\cite{Brahmbhatt_2020_ECCV, zhang2018learning}, or visual understanding~\cite{Shiota_2024_WACV, Damen2022RESCALING,darkhalil2022epic}. Our work supports all of these tasks by providing a high quality model and evaluations.

\newcommand{\citeh}[0]{\textcolor{red}{\bf [cite here]}\xspace}
\section{Method}

The goal of this approach is to take an image $\mathcal{I}$ and jointly detect any visible \textcolor{myred}{hands},  \textcolor{myyellow}{1st object} and \textcolor{mycyan}{2nd object} and predict
their pairwise interaction relationships. We build on the state-of-the-art Co-DETR architecture~\cite{zong2023detrs}, which offers a strong and general object detector, and show that training it for this task
with a lightweight interaction module, end-to-end, can perform detection and interaction prediction efficiently, in a single forward pass.

\begin{figure*}[t]
    \centering
    \includegraphics[width=1.0\linewidth]{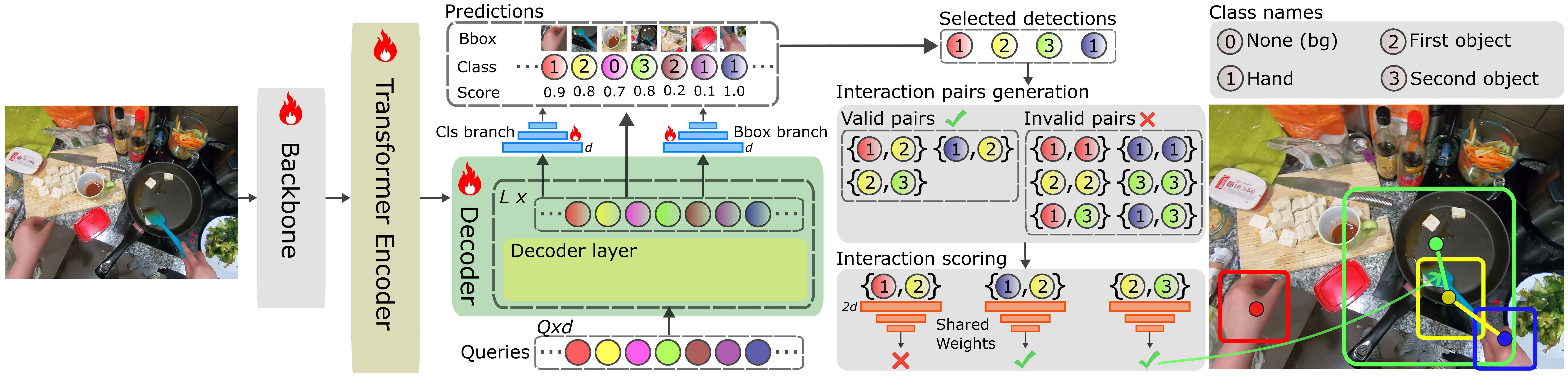}
    \vspace*{-12pt}
    \caption{\textbf{HOI-DETR.}
    We train a Co-DETR backbone end-to-end to predict hands, objects and their interactions. 
    We add an interaction MLP head to the decoder outputs to predict pairwise contact relations between hands, objects and within objects.
    We use the class predictions head to decide possible pairs for interactions (hand-object) and (object-object). 
    The module operates on contextualised token embeddings and produces relation probabilities.
    We visualise predicted bounding boxes, their classes and interaction links.}    
    \label{fig:method}
    \vspace*{-8pt}
\end{figure*}

Figure~\ref{fig:method} shows an overview of the complete method. There are three primary components: (1) a transformer backbone and encoder that map from $\mathcal{I}$ to a set of latent feature tokens; (2) a decoder that maps $Q$ query tokens to detections (a bounding box, class, and score) while cross-attending to the latent feature tokens; and (3) an interaction module that classifies the interaction relationship of a pair of embeddings from the decoder representing boxes. 
During training, Co-DETR has extensive auxiliary supervision, including additional heads that are used {\it only} at training time. We sketch these heads and losses for completeness, but refer the reader to~\cite{zong2023detrs} for a complete description. 

\subsection{Backbone and Transformer Encoder}

The backbone and transformer encoder map an input image $\mathcal{I}$ to a set of $T$ latent tokens. %
The backbone is a 
ViT~\cite{dosovitskiy2021an} that converts a patchified image into a sequence of patch embeddings.
These embeddings, plus positional encodings, are passed into a transformer encoder, creating the final $T$ latent tokens.

Following Co-DETR, the encoder processes the multi-scale backbone features with
multi-scale deformable attention, producing a refined latent representation
\(
F \in \mathbb{R}^{H \times W \times D}.
\)
This latent feature map is then adapted into a multi-scale feature
pyramid
\(
\{F_1,\ldots,F_J\}
\)
via the multi-scale adapter. Both $F$ and $\{F_j\}$
provide the feature
maps used by the auxiliary heads.

\vspace{1mm}
\par \noindent 
{\it Auxiliary encoder heads.} Co-DETR trains $K$ auxiliary heads on multi-scale copies of the features to minimize detection losses~\cite{ren2015faster,zhang2020bridging}.
All spatial locations of the feature maps $\{F_j\}$ act as candidate points for
one-to-many assignments. A ground-truth box may match multiple encoder
locations, forming a positive set
\(
\mathcal{P} = \{p : p \leftrightarrow \text{GT}\}.
\)
Positive locations receive both classification and regression losses; all
unmatched/negative locations contribute to the classification loss only:
\begin{equation}
\mathcal{L}^{enc}
=
\sum_{k=1}^{K}
\Big[
\mathcal{L}^{enc}_{cls,k}(\text{pos+neg})
+
\mathcal{L}^{enc}_{reg,k}(\text{pos only})
\Big].
\label{eq:enc_loss}
\end{equation}
This versatile one-to-many supervision, also used by \cite{shan2020understanding, cheng2023towards},
enriches the encoder’s spatial discrimination and stabilises training.

\subsection{Transformer Decoder}
\label{subsec:decoder}
The transformer decoder cross-attends to the $T$ latent image tokens to convert a set of $Q$ fixed $d$-dimensional object queries to a set of $Q$ detections across $C$ classes. 
Different from Co-DETR, which performs semantic category detection, our decoder predicts role-based interaction classes defined as 
$\{0$:~\textcolor{myred}{hands}, 
$1$:~\textcolor{myyellow}{1st object}, 
$2$:~\textcolor{mycyan}{2nd object}, 
$3$:~\textcolor{mygray}{background}$\}$. 
These labels are based on their role in the particular image rather than their category and the same object may change class depending on its relationship to a hand (as shown in Figure~\ref{fig:back12example}).

The decoder consists of $L$ layers ($L=6$), each of which performs self-attention across the queries followed by cross-attention on the encoder features. At the decoder layer $l$, for each query $i$, the transformer has a feature $\hB^{l}_i$ as well as logit prediction $\cB_i^{l} \in \mathbb{R}^{C+1}$ and normalised bounding box prediction $\bB_i^{l} \in [0,1]^4$. The final prediction of the decoder is the last layer's predictions: $\bB_i^{L}, \cB_i^{L}$.

At each layer $l$, the decoder is supervised in a standard DETR style, where the $Q$ box predictions are matched with Hungarian matching~\cite{kuhn1955hungarian} to the $N$ ground-truth instances in order to minimize the total loss. The final loss, given the matches, follows Co-DETR and is the sum of a classification loss, L1 bounding-box regression, and GIOU.
The corresponding decoder loss at this stage is denoted as $\hat{\mathcal{L}}^{\text{dec}}_{l}$, representing the loss for layer $l$ in the $main$ one-to-one branch.

\vspace{1mm}
\par \noindent {\it Auxiliary losses.} To help training, Co-DETR trains the decoder (sharing weights but operating separately) on another set of queries. These queries are generated dynamically from the auxiliary dense prediction head results from the backbone and encoder~\cite{zong2023detrs}.
An added term $\mathcal{L}^{\text{dec}}_{i,l}$ is used for the $auxiliary$ one-to-many branches, where $i$ is the auxiliary head index.

\subsection{Interaction Module}

To predict interactions between hands and objects, we add an additional head to Co-DETR in order to identify whether a pair of predicted bounding boxes (i.e., objects) $i$ and $j$ have an interaction relationship.
As the latent in the decoder encodes information about the class, the spatial position (as it is used to regress the bounding box) and the detection confidence, it is best positioned as the input to this prediction head.
The proposed interaction head maps a pair of decoder latents, at each layer $l$, ($\hB_i^{l}$  and $\hB_{j}^{l}$) to a binary classification. Specifically, for each pair $i,j$ we apply a MLP to the concatenated latents
$[\hB_i^{l} || \hB_j^{l}] \in \mathbb{R}^d$ to produce logits $\zB_{i,j}^{l}$. Importantly, the pair is ordered, so if two objects are predicted as interacting, $h_i^{l}$ would be the first object and $h_j^{l}$ would be the second object. 

At each layer $l$, the logits from the model are supervised for valid pairs that could have an interaction. If $y_i$ is the $i$th box's label, the only valid pairs are:
\begin{equation}
\mathcal{P} = 
\{(i,j)~~\text{s.t.}~~(y_i,y_j) \in \{(\textcolor{myred}{\text{hand}},\,\textcolor{myyellow}{\text{1st object}}),
(\textcolor{myyellow}{\text{1st object}},\,\textcolor{mycyan}{\text{2nd object}})\}\}.
\end{equation}
We exclude \textcolor{myred}{hand}~$\to$~\textcolor{mycyan}{2nd object} pairs, as by definition a hand can only be interacted with \textcolor{mycyan}{\text{2nd object}} through \textcolor{myyellow}{\text{1st object}}. Each pair $(i,j)$ in $\mathcal{P}$ is given a label $r_{ij}$ according to the ground-truth interaction, and the interaction module's predicted logit $z_{ij}^l$ is supervised at every layer of the decoder with a focal loss. If $\text{FL}$ is the focal loss~\cite{lin2017focal},
\begin{equation}
\mathcal{L}^{int}_{l} = 
\frac{1}{|\mathcal{P}|} \sum_{(i,j) \in \mathcal{P}} \text{FL}(z_{i,j}^l, r_{i,j}).
\end{equation}

\begin{figure}[t]
    \centering
    \includegraphics[width=1.0\linewidth]{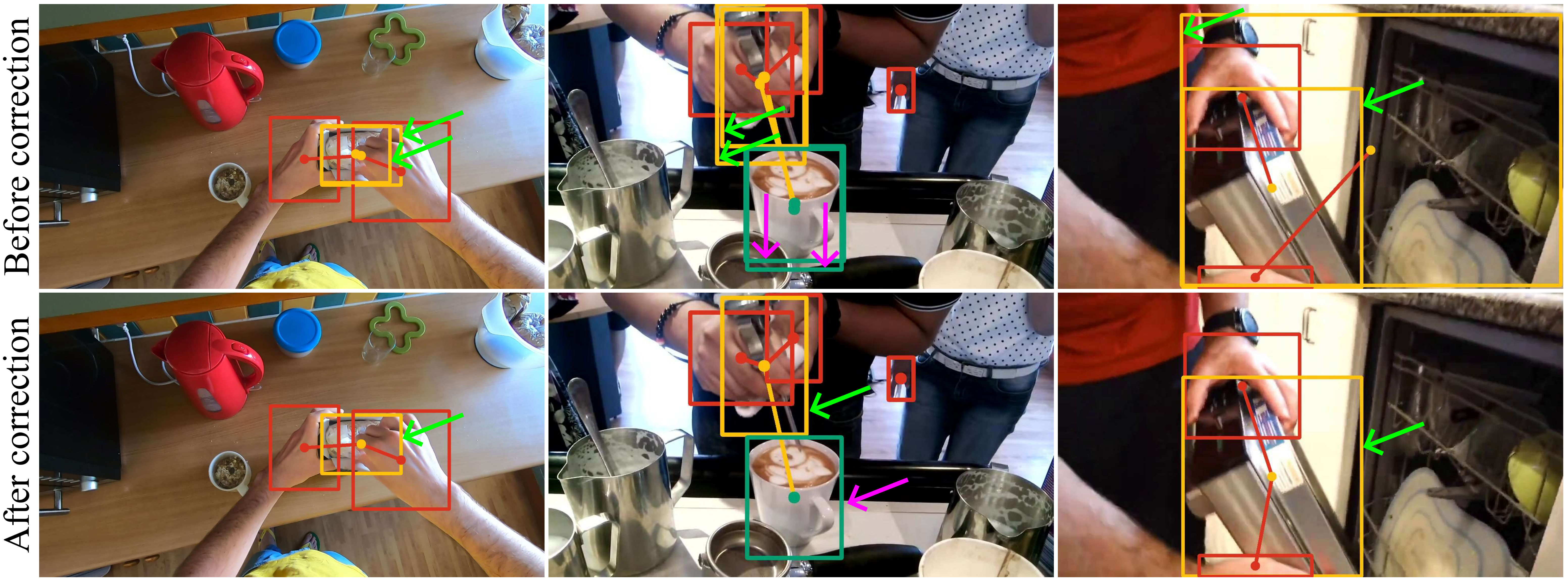}
    \vspace*{-14pt}
    \caption{Visualisation of Hands23 dataset before and after correction.
    Duplicate boxes were removed and all hand--object links were remapped to ensure consistency. The arrows highlight the bounding boxes that were corrected.
    } 
    \vspace{-8pt}
    \label{fig:hands_correction}

\end{figure}

We combine the  interaction loss with the standard Co-DETR losses, yielding:
\begin{equation}
\sum_{l=1}^{L}
\Big(
\hat{\mathcal{L}}^{dec}_{l}
+
\lambda_{1}\sum_{i=1}^{K}\mathcal{L}^{dec}_{i,l}
+
\lambda_{2}\mathcal{L}^{int}_{l}
\Big) +\lambda_{3}\,\mathcal{L}^{enc}.
\label{eq:global_loss}
\end{equation}
Here, $\hat{\mathcal{L}}^{dec}_{l}$ and $\mathcal{L}^{dec}_{i,l}$ denote the layer-wise losses from the main one-to-one decoder branch and the auxiliary one-to-many branches, respectively. $\mathcal{L}^{enc}$ is the total encoder supervision loss, and $\mathcal{L}^{int}_{l}$ is the proposed interaction loss. The weights $\lambda_{1}$ and $\lambda_{3}$ follow Co-DETR’s  settings, and $\lambda_{2}$ controls the strength of the proposed interaction supervision. We ablate $\lambda_{2}$  in the appendix.

\subsection{Inference}

At test time, we follow Co-DETR. We map the image to latent tokens, followed by detection with the decoder. The decoder's predictions are processed with  Co-DETR post-processing on the 
\emph{final} decoder layer ($l{=}L$):
softmax scoring, greedy thresholding, and multi-class NMS. We then apply the interaction module {\it only} to surviving detections and valid pairs, resulting in directed relationships between detected
\textcolor{myred}{hands} ${\rightarrow}$
\textcolor{myyellow}{1st objects},
and \textcolor{myyellow}{1st objects} $\rightarrow$
\textcolor{mycyan}{2nd objects}.

\section{Datasets}
\label{sec:datasets}

\begin{figure}[t]
    \centering    \includegraphics[width=\textwidth]{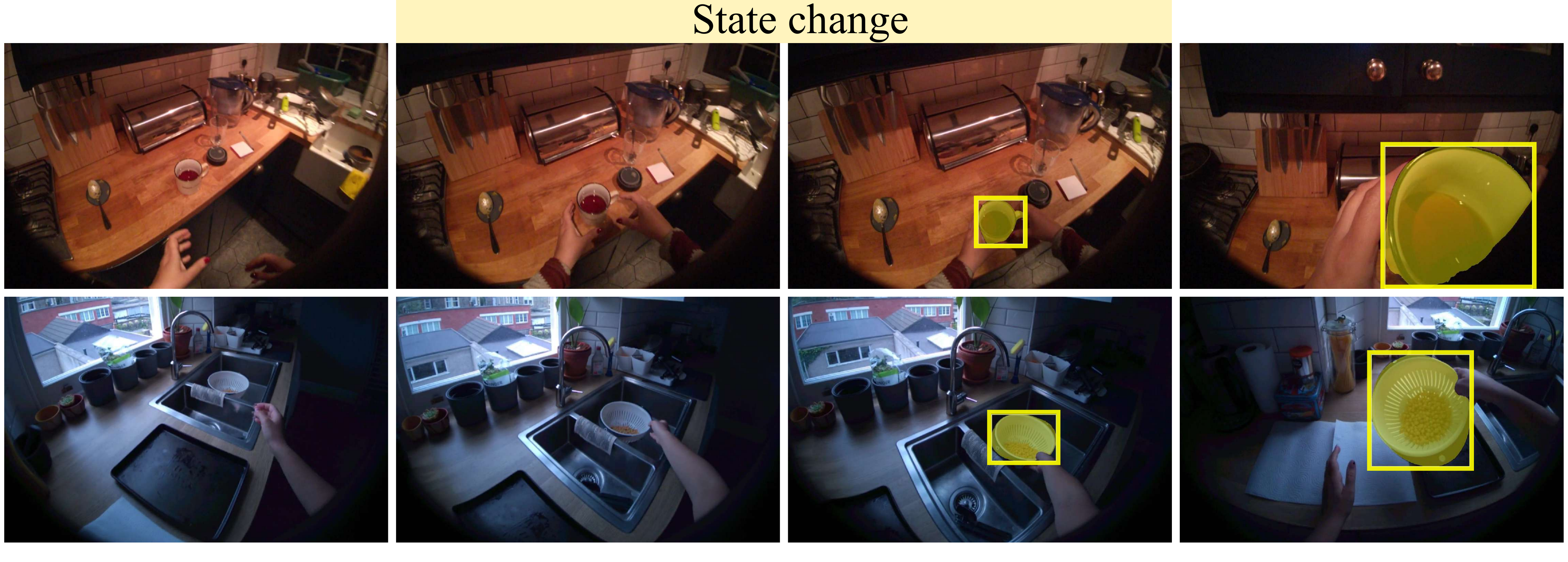}
    \vspace{-4.0ex}
    \caption{Visualisation of HD-EPIC-HOI dataset.
    We generate pseudo labels of the \textcolor{myyellow}{1st object} around the contact state change for the tracks where only one object is moving.
    For simplicity, the figure shows four sparsely sampled frames out of 1.5 seconds sequences as explained in section \ref{sec:datasets}.} \vspace*{-18pt}
    \label{fig:hd_epic_examples}

\end{figure}

\vspace*{-6pt}

\begin{wraptable}{r}{0.57\textwidth}
    \centering
    \vspace{-1.5ex}
    \resizebox{\linewidth}{!}{%
    \begin{tabular}{lcccc}
        \toprule
        Dataset & Images & \textcolor{myred}{Hand} & \textcolor{myyellow}{1st obj} & \textcolor{mycyan}{2nd obj} \\
        \midrule
        Hands23~\cite{cheng2023towards} & 24.6k & 39.8k & 2.5k & 2.5k \\
        HD-EPIC-HOI~\cite{perrett2025hdepic} & 41.9k & -- & 26k & -- \\
        HOIST~\cite{sn_hoist_cvpr_2024} & 3.5k & -- & 3.9k & -- \\
        FineBio~\cite{yagi2025finebio} & 238 & 465 & 372 & -- \\
        \bottomrule
    \end{tabular}%
    }
    \vspace{-1.5ex}
    \caption{ Evaluation splits. ``--'' indicates no annotations are available. Hands23 is our refined version.}
    \vspace{-1ex}
    \label{tab:dataset_stats}
\end{wraptable}
One key ingredient for progress in HOI has been datasets for training and evaluation. In this work, we use several datasets for evaluation, including 
Hands23~\cite{cheng2023towards},
and HD-EPIC~\cite{perrett2025hdepic},
HOIST~\cite{sn_hoist_cvpr_2024} and FineBio~\cite{yagi2025finebio}, as shown in Table~\ref{tab:dataset_stats}.
These datasets are chosen to span a variety of tasks and settings. 
In some cases, these can be used directly. HOIST~\cite{sn_hoist_cvpr_2024} focuses on providing denser video annotations of hand-held objects only, which aligns with our \textcolor{myyellow}{1st object} definition. FineBio~\cite{yagi2025finebio} is a test of generalisation, as it depicts biology experiments from an egocentric perspective and is therefore substantially different than any of the models' training data. It offers annotations for hands and first objects.
In two cases, we provide new
data that helps improve and measure performance of hand-detection systems. 
First, we contribute annotation refinements to Hands23~\cite{cheng2023towards} that improve our ability to predict HOI and measure its performance. Second, we use HD-EPIC~\cite{perrett2025hdepic} to produce an evaluation set, named HD-EPIC-HOI that helps measure performance.

\begin{figure}[t]
    \centering
    \includegraphics[width=\linewidth]{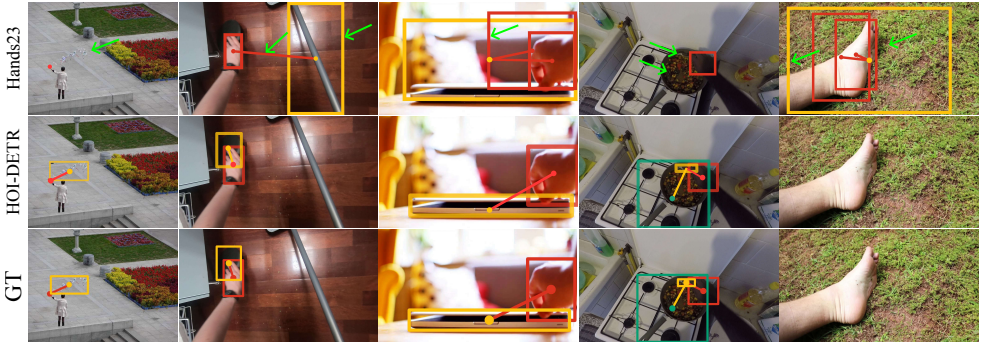}
    \vspace*{-16pt}
    \caption{Qualitative results of HOI-DETR and Hands23 model on Hands23 dataset. Arrows are included to highlight key errors.}
    \vspace{-2.0ex}
    \label{fig:qualititve_image_level1}
\end{figure}

\noindent\textbf{Refined Hands23}. Hands23~\cite{cheng2023towards} introduced a new large dataset consisting of 257K images, 401K hands, 288K objects, and 19K second objects. One challenge in Hands23 is that its annotation pipeline resulted in duplicated object annotations. These stem from a hand-centric scheme, where each hand was separately annotated. If hands are engaged in bimanual interaction, then two bounding boxes are independently, and manually, annotated for the same physical object. This is most obvious when the results are  contradictory -- i.e. the object extent is ambiguous (e.g., when a hand opens a fridge, either the door or the full fridge might be boxed).
Following an initial round of training, we realised this annotation error impacts both the training loss and the evaluation metrics.

To address this issue, we built a refinement pipeline that removes duplications. For each train/val image where annotated objects have the same category (\textcolor{myyellow}{first}/\textcolor{mycyan}{second} object) and some overlap (IoU above a threshold), annotators label whether the two boxes are the same physical object. 
We manually annotated 26.2k images across the training and validation splits; of these, 56.2\% was corrected. In the training set, 10{,}730 of 20{,}976 images with IoU overlap (51.2\%) were corrected, while in the validation set, 4{,}135 of 5{,}529 images (74.8\%) required corrections. These images yielded 31k  bounding-box pairs, of which 16{,}208 were labeled as \textit{same object} and 14{,}788 as distinct instances. 
We maintained one of the two duplicate bounding boxes for the same object, and the interaction links were remapped accordingly to maintain consistency.
\Cref{fig:hands_correction} shows correction samples, which shows the removal of inconsistent or duplicated boxes.

\noindent\textbf{HD-EPIC-HOI}.
We introduce a new video benchmark for HOI detection, derived from the new HD-EPIC dataset~\cite{perrett2025hdepic}. HD-EPIC provides high-resolution egocentric videos with temporally localised segments of object motion -- spanning the duration from when an object is first picked up until it is placed back down.

HD-EPIC-HOI annotations are particularly interesting for HOI detection for two reasons. First, frames near interactions (i.e., contact and release) are significantly more challenging for an HOI detector than random frames.
This is because when the hand is approaching (or releasing) an object, those \emph{near contact} frames are more likely to be mis-detected as contact (directly before the contact) compared to frames where hands are clearly far from any objects.
Second, the temporal annotations allow calculating spatio-temporal measures of HOI consistency, even when detections are frame-level.

We sample 1 second ($\sim$30 frames) evaluation sequences from HD-EPIC centered around the \textit{start} or the \textit{end} of an object motion.
We propagate the sparse manual ground-truth masks of the moving object to all frames in the sequence with Segment Anything 2 (SAM2)~\cite{ravi2025sam} followed by a human check of the generated sequences, to discard sequences with annotation noise.
This produces temporally coherent object masks and  boxes.
This dataset can be used for evaluating \textcolor{myyellow}{1st object} detections, frame-wise and temporally.

HD-EPIC-HOI consists of manually verified 768 sequences with 35.3k images and 22.2k \textcolor{myyellow}{1st object} bounding boxes.
\Cref{fig:hd_epic_examples} shows several examples from HD-EPIC-HOI tracks.

\section{Experiments}
\label{sec:exps}

Our experiments test the effectiveness of HOI-DETR across multiple datasets.
We test HOI-DETR's frame-level accuracy on four datasets, including  cross-domain experiments, using standard metrics. For spatiotemporal consistency, we use two metrics and test on clips from our new HD-EPIC-HOI dataset. 

\noindent \textbf{Implementation Details.}
Throughout, we use a 24-layer ViT-L/16 with 16 attention heads and window size of 24 and a 6-layer transformer encoder.
We employ Focal Loss ($\alpha{=}0.25$, $\gamma{=}2.0$) for interaction classification and use the following detection loss weights: L1=5 and GIoU=2.
Training uses the \texttt{AdamW} optimiser with gradient clipping ($\text{max\_norm}=0.1$) and a 500-iteration linear warm-up.
We initialise HOI-DETR from a Co-DETR model pretrained on COCO~\cite{lin2014microsoft} and fine-tune it on the refined Hands23 dataset for 5 epochs, using a batch size of 16 and multi-scale augmentation centered at a resolution of $640 \times 640$.
The loss weights are set to $(\lambda_{1}, \lambda_{2}, \lambda_{3}) = (1.0,\, 3.0,\, 2.0)$.

\noindent\textbf{Frame-level Metrics.} We evaluate hand/object detection and interaction:
\begin{itemize}[topsep=0pt, itemsep=0pt, partopsep=0pt, parsep=0pt, leftmargin=*]
\item[$\bullet$] {\it \textcolor{myred}{Hand}/\textcolor{myyellow}{1st object}/\textcolor{mycyan}{2nd object Detection}}: We follow standard practice and use mAP at 50\% IoU as our primary metric; the looser  IoU requirement compared to the COCO~\cite{lin2014microsoft} protocol follows conventions in hand-detection literature~\cite{shan2020understanding} that stem from difficulty precisely annotating the wrist. 
\item[$\bullet$] {\it Interaction Prediction}: We use the F1 score following~\cite{cheng2023towards}, as it provides a balanced measure between precision and recall.
Positive queries are determined via Hungarian matching with the ground-truth annotations. %
\end{itemize}

\begin{figure}[t]
    \centering
    \includegraphics[width=\linewidth]{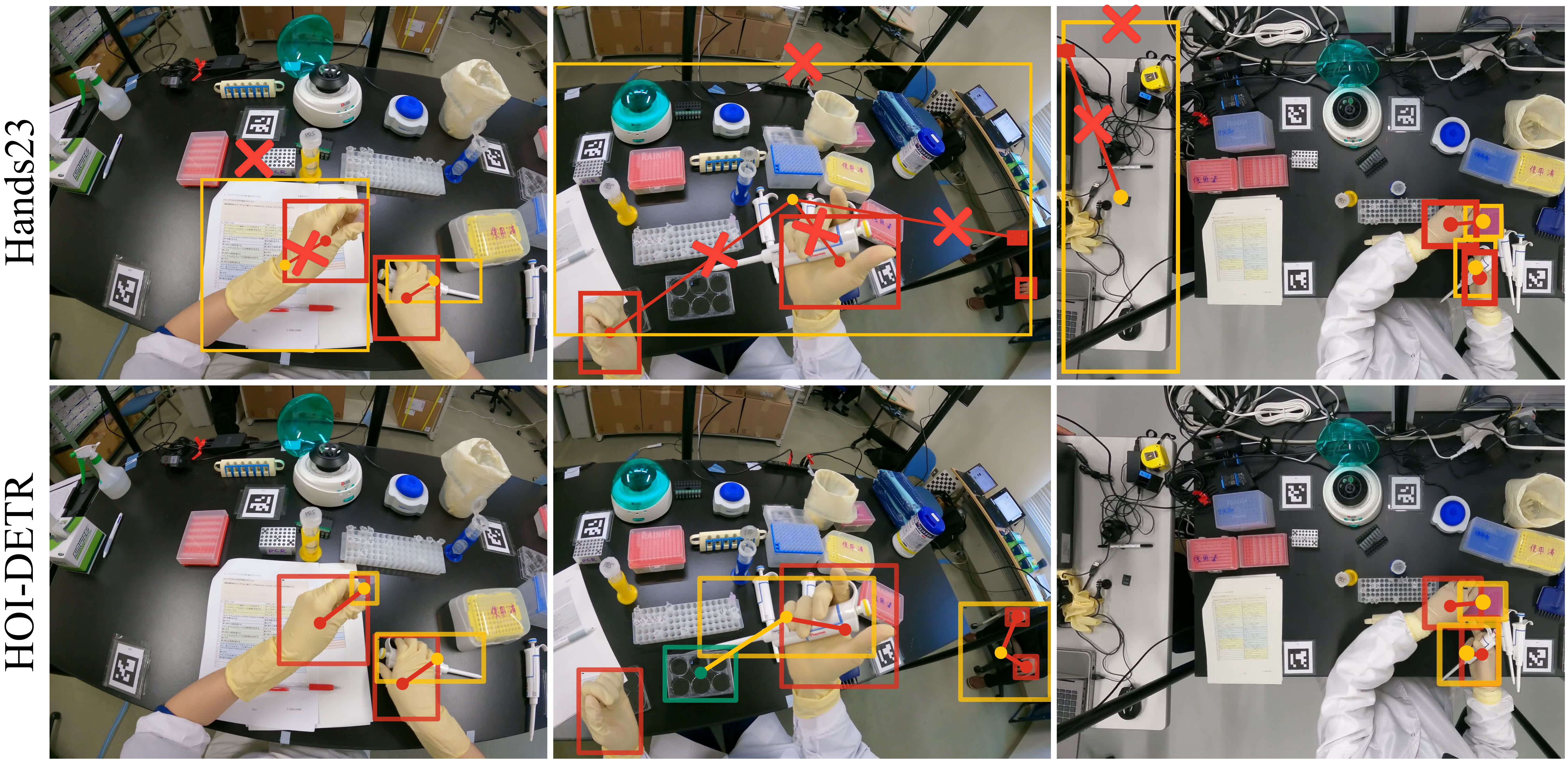}
    \vspace{-4ex}
    \caption{
        \text{Qualitative results on the FineBio~\cite{yagi2025finebio} dataset.}
        We compare Hands23 and HOI-DETR.
        ``\textcolor{red}{\ding{55}}'' indicates an \emph{invalid interaction link} or a \emph{false-positive} prediction produced by a model. HOI-DETR handles \emph{out-of-distribution} footage.
    } 
    \label{fig:qualitative_finebio}
\end{figure}

\begin{figure}[t]
    \centering
    \includegraphics[width=\linewidth]{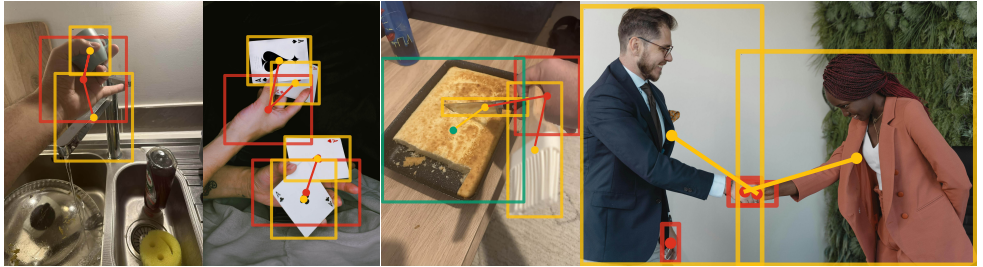}
    \vspace{-4ex}
    \caption{\text{Complex interactions.} HOI-DETR predicts independent pairwise interaction scores.
    It can thus detect one-to-many relationships where one hand is interacting with multiple objects and hand-to-hand interactions. Note that hand shaking (right) is modelled as the hand interacting with the full body of the other person.}
    \vspace*{-16pt}\label{fig:hand_to_multiple_objs}
\end{figure}

\noindent\textbf{Spatiotemporal Consistency Metrics.}
We evaluate the spatiotemporal consistency with two metrics. These are important because
intermittent flickering impacts downstream tasks but has negligible impact on frame-averaged mAP. %
\begin{itemize}[topsep=0pt, itemsep=0pt, partopsep=0pt, parsep=0pt, leftmargin=*]
\item[$\bullet$] {\it Video-level AP (Video-AP)}. 
Following~\cite{kalogeiton2017action, li2020actions}, we evaluate spatio-temporal consistency using Video-AP, which extends 
frame-level AP to the video level by treating predicted bounding box trajectories as tubes 
and matching them against ground-truth tubes via spatio-temporal IoU.

\item[$\bullet$] {\it Long-Term Temporal Consistency (LTC)}.
LTC quantifies the maximum uninterrupted duration for which a model continuously follows the correct object before losing it. 
A frame is considered successfully tracked if its best-matched prediction satisfies $\mathrm{IoU} \ge 0.5$. For each video, we collect the set of continuously tracked trajectories~$\mathcal{T}$. If the length of a trajectory $\tau \in \mathcal{T}$ is denoted $|\tau|$, the LTC for a video is defined as the maximum trajectory length, normalised by the video length $F$, or $(1/F) \max_{\tau \in \mathcal{T}} |\tau|$.
We report the average LTC; the $1/F$ ensures that long videos do not overly influence average LTC over a dataset.
\end{itemize}

\noindent\textbf{Baselines.} Our main baseline is the Hands23 detector~\cite{cheng2023towards}, which showed better performance compared to the 100-DOH~\cite{shan2020understanding} detector. This method extends Mask-RCNN to perform a number of hand-related tasks. For video datasets, we additionally compare with HOIST~\cite{sn_hoist_cvpr_2024}, a transformer-based model that uses multiple frames. We obtain boxes for HOIST by converting the predicted masks into bounding boxes.
Finally, to evaluate pure hand detection capabilities, we compare against general pose models and specialised hand detectors including MediaPipe~\cite{zhang2020mediapipe}, OpenPose~\cite{cao2017realtime}, ContactHands~\cite{narasimhaswamy2019contextual}, ViTDet~\cite{li2022exploring}, and  WiLOR~\cite{potamias2025wilor}.

\noindent\textbf{Frame-Level Quantitative Results.} We report results on  Hands23~\cite{cheng2023towards} (Table~\ref{tab:results_hands23_combined}), HD-EPIC-HOI (Table~\ref{tab:results_hdepic_consolidated}), FineBio~\cite{yagi2025finebio} (Table~\ref{tab:results_finebio}) and HOIST (Table~\ref{tab:results_hoist}). In all cases, HOI-DETR shows a large improvement over the state of the art, even when on unseen datasets (e.g., HOIST) or domains (e.g., FineBio). First object gains are particularly large compared to Hands23 ($+25\%$ absolute increase on multiple datasets), and interaction error rate is reduced by $52\%$. Moreover, despite detecting frame-by-frame, HOI-DETR outperforms the video-based HOIST, and generalises far better ($72.6$ vs $30.4$) to the new HD-EPIC-HOI dataset. More frame-level evaluations including per-size and per-subset is provided in the supplementary.

\begin{table*}[!t]
\centering

\resizebox{\textwidth}{!}{%
\begin{tabular}{llccccccccc}
\toprule
\multirow{2}{*}{Method} & \multirow{2}{*}{Training Data}
& \multicolumn{4}{c}{\textbf{Evaluation: Original~\cite{cheng2023towards}}} 
& \multicolumn{5}{c}{\textbf{Evaluation: Refined}} \\
\cmidrule(lr){3-6} \cmidrule(lr){7-11}
& & Overall & \textcolor{myred}{Hand} & \textcolor{myyellow}{1st obj} & \textcolor{mycyan}{2nd obj}
& Overall & \textcolor{myred}{Hand} & \textcolor{myyellow}{1st obj} & \textcolor{mycyan}{2nd obj} & $\text{F1}_{\text{inter}}$\\
\midrule
Hands23~\cite{cheng2023towards} 
& Original
& 60.2 & 85.2 & 51.6 & 43.8 
& 63.6 & 85.2 & 59.4 & 46.2 & 90.7 \\

HOI-DETR (Ours)
& Original
& \textbf{82.3} & \textbf{93.1} & \textbf{78.6} & \textbf{75.3}
& 86.0 & \textbf{93.1} & 86.3 & 78.5 & 91.4 \\

HOI-DETR (Ours)
& Refined
& 81.9 & \textbf{93.1} & 77.4 & 75.2
& \textbf{86.1} & \textbf{93.1} & \textbf{86.5} & \textbf{78.7} & \textbf{95.5} \\
\bottomrule
\end{tabular}%
}
\vspace{-1.5ex}
\caption{Comparison on Hands23 val set (Detection AP$_{50}$) training on original and refined data. HOI-DETR substantially improves over the Hands23~\cite{cheng2023towards} detector; 1st/2nd object performance is consistently higher on the refined data due to removal of duplicate objects.}
\label{tab:results_hands23_combined}
\end{table*}

\begin{figure}[t]
    \centering
    \includegraphics[width=\linewidth]{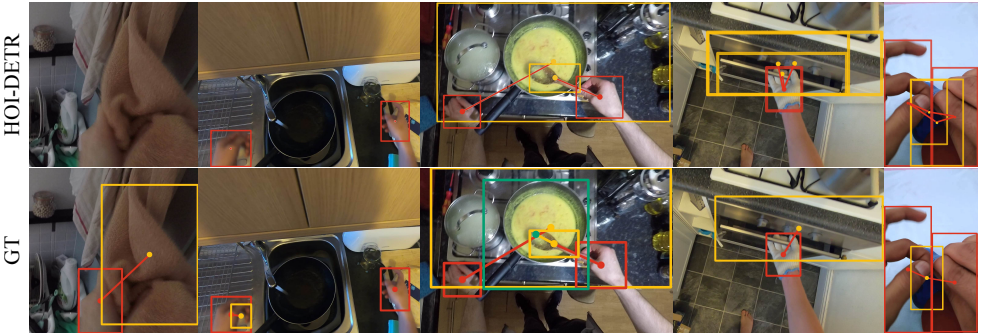}
    \caption{
        Challenging failure cases from Hands23 dataset, comparing HOI-DETR predictions against the ground truth (GT). 
    }
    \label{fig:qualitative_image_level_random_with_failures}
    \vspace{-12pt}
\end{figure}

\begin{figure}[t]
    \centering
    \label{fig:qualitive_video}
    \includegraphics[width=\linewidth]{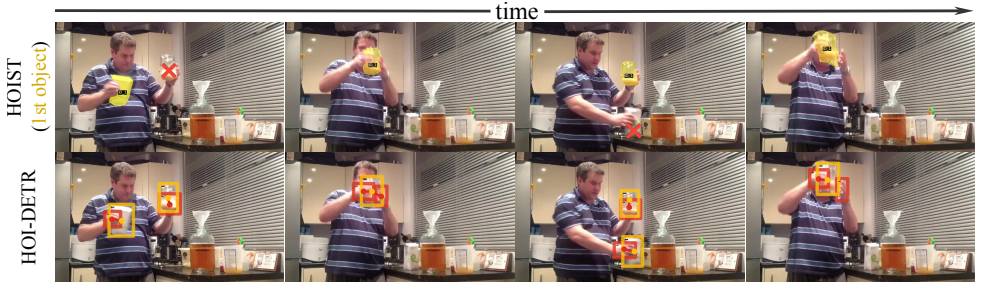}
    \vspace*{-16pt}
    \caption{\text{Qualitative video-sequence from HOIST dataset.} We compare HOI-DETR with HOIST model. HOIST predicts \textcolor{myyellow}{1st object} only, ``\textcolor{red}{\ding{55}}'' indicates the \emph{errors}.
    }
    \label{fig:qualitive_video_hoist}
    \vspace*{-6pt}
\end{figure}

\begin{figure}[t]
    \centering
    \includegraphics[width=\linewidth]{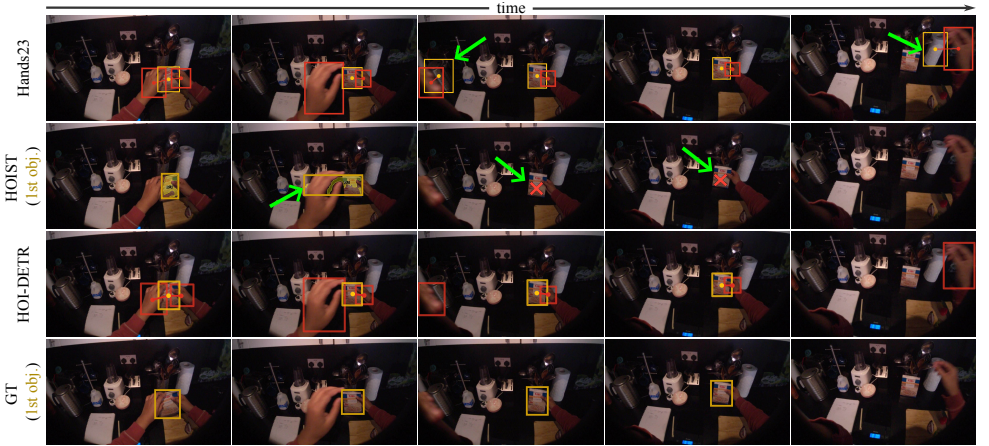}
    \caption{\text{Qualitative video-sequence from HD-EPIC-HOI.} We compare Hands23, HOIST and HOI-DETR. The ground truth (GT) annotations provide only the \textcolor{myyellow}{1st object} annotations, but we also show all predicted hand regions for both Hands23 and HOI-DETR. \textcolor{green}{Green} arrows are included to highlight key errors.
    }
    \label{fig:qualitive_video}
    \vspace{-12pt}
\end{figure}

\begin{table}[t]
\centering
\begin{tabular}{lcccccc}
\toprule
Method & Frame-AP & $\Delta$ & Video-AP & $\Delta$ & LTC & $\Delta$ \\
\midrule
Hands23~\cite{cheng2023towards} & 46.9 & \textcolor{red}{-25.7} & 26.8 & \textcolor{red}{-33.4} & 31.4 & \textcolor{red}{-29.6} \\
HOIST~\cite{sn_hoist_cvpr_2024} & 30.4 & \textcolor{red}{-42.2} & 16.1 & \textcolor{red}{-44.1} & 27.2 & \textcolor{red}{-33.8} \\

HOI-DETR (Ours) & \textbf{72.6} & -- & \textbf{60.2} & -- & \textbf{61.0} & -- \\
\bottomrule
\end{tabular}
\vspace{1ex}
\caption{\text{Comprehensive Performance on HD-EPIC-HOI.} We evaluate frame-level detection (Frame-AP), spatiotemporal consistency (Video-AP), and tracking duration (LTC). $\Delta$ denotes the absolute performance drop of prior methods compared to our proposed approach. The baselines suffer massive deficits across all metrics, particularly in temporal consistency.}
\label{tab:results_hdepic_consolidated}
\end{table}

\begin{table}[t]
\centering
\small 
\setlength{\tabcolsep}{3.5pt} %

\begin{minipage}[t]{0.3\textwidth} 
\centering
\begin{tabular}[t]{lcc}
\toprule
Method & HOIST & $\Delta$ \\
\midrule
Hands23~\cite{cheng2023towards} & 43.1 & \textcolor{red}{-33.5} \\
HOIST~\cite{sn_hoist_cvpr_2024} & 70.7 & \textcolor{red}{-5.9} \\
\midrule
HOI-DETR & \textbf{76.6} & -- \\
\bottomrule
\end{tabular}
\vspace{-2ex}
\caption{1st obj AP$_{50}$ for HOIST~\cite{sn_hoist_cvpr_2024}.}
\vspace*{-20pt}
\label{tab:results_hoist}
\end{minipage}
\hfill
\begin{minipage}[t]{0.6\textwidth} %
\centering
\begin{tabular}[t]{lcccccc}
\toprule
Method & Overall & $\Delta$ & \textcolor{myred}{Hand} & $\Delta$ & \textcolor{myyellow}{1st obj} & $\Delta$ \\
\midrule
Hands23~\cite{cheng2023towards} & 50.2 & \textcolor{red}{-21.0} & 74.4 & \textcolor{red}{-12.2} & 26.0 & \textcolor{red}{-29.8} \\
 & & & & & & \\ 
\midrule
HOI-DETR  & \textbf{71.2} & -- & \textbf{86.6} & -- & \textbf{55.8} & -- \\
\bottomrule
\end{tabular}
\vspace{-2ex}
\caption{AP$_{50}$ for FineBio~\cite{yagi2025finebio} Val. $\Delta$ denotes the absolute performance deficit compared to our method.}
\label{tab:results_finebio}
\end{minipage}
\vspace{-2.0ex}
\end{table}

\definecolor{zeroshot}{RGB}{208,226,255}
\begin{table*}[t]
\centering
\resizebox{\textwidth}{!}{%
\begin{tabular}{l c c c c c c c} 
\toprule
Method & Hand+Obj & Hands23 & WHIM & COCO-Whole & Oxford-Hands & FineBio & EgoHands \\
\midrule
MediaPipe~\cite{zhang2020mediapipe} & $\times$ & -- & \cellcolor{zeroshot} $^*$53.1 & \cellcolor{zeroshot} $^*$15.4 & \cellcolor{zeroshot} $^*$8.7 & -- & -- \\
OpenPose~\cite{cao2017realtime} & $\times$ & -- & \cellcolor{zeroshot} $^*$76.8 & \cellcolor{zeroshot} $^*$37.1 & \cellcolor{zeroshot} $^*$20.7 & -- & -- \\
ContactHands~\cite{narasimhaswamy2019contextual}& $\times$ & -- & \cellcolor{zeroshot} $^*$\underline{93.4} & \cellcolor{zeroshot} $^*$50.3 & \cellcolor{zeroshot} $^*$70.0 & -- & -- \\
ViTDet~\cite{li2022exploring} & $\times$ & -- & \cellcolor{zeroshot} $^*$84.7 & \cellcolor{zeroshot} $^*$41.6 & \cellcolor{zeroshot} $^*$67.6 & -- & -- \\
WiLOR~\cite{potamias2025wilor} & $\times$ & \cellcolor{zeroshot} 77.3 & $^*$\textbf{96.1} & \cellcolor{zeroshot} $^*$62.5 & \cellcolor{zeroshot} $^*$\textbf{82.6} & \cellcolor{zeroshot} \underline{76.5} & \cellcolor{zeroshot} \underline{93.4} \\
\midrule
Hands23~\cite{cheng2023towards} & $\checkmark$ & \underline{85.2} & \cellcolor{zeroshot} \phantom{$^*$}71.7 & \cellcolor{zeroshot} \phantom{$^*$}\underline{67.0} & \cellcolor{zeroshot} \phantom{$^*$}60.6 & \cellcolor{zeroshot} 74.4 & \cellcolor{zeroshot} 93.0 \\
HOI-DETR (Ours) & $\checkmark$ & \textbf{93.1} & \cellcolor{zeroshot} \phantom{$^*$}78.1 & \cellcolor{zeroshot} \phantom{$^*$}\textbf{75.0} & \cellcolor{zeroshot} \phantom{$^*$}\underline{74.0} & \cellcolor{zeroshot} \textbf{86.6} & \cellcolor{zeroshot} \textbf{98.5} \\
\bottomrule
\end{tabular}
}
\vspace{-1ex} 
\caption{Hand detection (AP$_{50}$) comparison. We compare HOI-DETR with specialised hand detectors and general pose models across multiple in- and out-of-domain benchmarks. The results show that HOI-DETR has robust zero-shot generalisation for pure hand detection across diverse datasets including WHIM~\cite{potamias2025wilor}, COCO-Whole~\cite{jin2020whole}, Oxford-Hands~\cite{mittal2011hand}, FineBio~\cite{yagi2025finebio} and EgoHands~\cite{urooj2018analysis}. 
\sethlcolor{zeroshot}
\hl{Zero shot combinations of a model and dataset are shown in light blue}. 
[$^*$: as reported in WiLOR~\cite{potamias2025wilor}.]}
\label{tab:hand_detection_generalization}
\end{table*}

\noindent\textbf{Frame-level Qualitative Results.}
We present qualitative comparisons between HOI-DETR and Hands23 in Figure~\ref{fig:qualititve_image_level1}. 
Hands23 often (i) misses the interacted \textcolor{myyellow}{1st object} (e.g., col.~1 and 2), (ii) confuses non-hand regions or other body parts with a \textcolor{myred}{hand} (e.g., cols.~3 and 5), and (iii) fails to detect both the \textcolor{myyellow}{1st} and \textcolor{mycyan}{2nd object} when they are small or cluttered (e.g., col.~4). In contrast, HOI-DETR correctly localises hands and target objects and predicts the intended interactions, demonstrating improved robustness to small objects, occlusion, and background clutter.
We further show results of the models on the FineBio~\cite{yagi2025finebio} dataset (Figure~\ref{fig:qualitative_finebio}), which is different from the training distribution (since the egocentric training data is primarily cooking). HOI-DETR produces a far more accurate understanding of the scene. 
More frame-level visualisations and challanges are provided in the supplementary.

\noindent\textbf{Complex Interactions Cases.}
Current benchmarks do not offer complex interaction scenarios involving multi-object in-hand, hand-to-hand interactions, and bimanual interactions. Because HOI-DETR predicts \emph{pairwise} interaction scores independently across all valid pairs, it naturally handles many-to-many relationships. In Figure~\ref{fig:hand_to_multiple_objs}, HOI-DETR successfully predicts a single hand interacting with multiple objects simultaneously (e.g., holding \textcolor{myyellow}{multiple cards}), bimanual interactions, and even hand-to-hand interactions such as handshaking (\textcolor{myred}{hand}$\rightarrow$ \textcolor{myyellow}{person}). 
While a comprehensive benchmark evaluating such cases does not currently exist, these qualitative results demonstrate the capabilities of HOI-DETR.

\noindent\textbf{Failure Cases.}
Examples in Figure~\ref{fig:qualitative_image_level_random_with_failures} highlight typical failure modes of HOI-DETR on Hands23 dataset.  
The HOI-DETR either confuses the \textcolor{myred}{hand} identity and \textcolor{myyellow}{1st object} due to colour similarity and lighting (e.g., first column),  misses \textcolor{myyellow}{1st object} for small objects with motion blur (e.g., second column left hand), misses the \textcolor{mycyan}{2nd object} (e.g., third column), or produces duplicate detections of the same physical object (e.g., last two columns).

\noindent\textbf{Spatiotemporal Consistency Quantitative Results.} 
We evaluate spatiotemporal consistency on our HD-EPIC-HOI benchmark (Table~\ref{tab:results_hdepic_consolidated}) by reporting the standard Video-AP and Long-Term Temporal Consistency (LTC). 
HOI-DETR achieves a Video-AP of $60.2$, outperforming Hands23 by $33.4$ points and HOIST by $44.1$ points, while also leading on LTC by a large margin ($61.0$ vs.\ $31.4$ and $27.2$). 
Notably, although HOIST is trained and evaluated on video sequences, it scores lower on Video-AP when compared to frame-based Hands23 and HOI-DETR. 
This significant performance gap underscores that video-level training alone does not guarantee temporally consistent detections. 
Our significant gains in both Video-AP and LTC validate the effectiveness of our training data and the superior representational capacity of the HOI-DETR architecture.

\noindent\textbf{Video-level Qualitative Results.} 
We present the video-level qualitative example in Figures~\ref{fig:qualitive_video_hoist} and~\ref{fig:qualitive_video}. The Hands23 method frequently produces false positives, most notably in the third and the last columns in Figure~\ref{fig:qualitive_video} where background regions are falsely detected as object with incorrect hand-object relation. HOIST method exhibits limited temporal consistency over both examples. In Figure~\ref{fig:qualitive_video}, it fails to detect the \textcolor{myyellow}{1st object} entirely, likely due to limited generalisation. In contrast, HOI-DETR demonstrates strong robustness across challenging frames, reliably detecting hands even when only small portions are visible (third column) or when motion blur is present (last column). 
Figure~\ref{fig:qualitive_video_hoist} shows that HOI-DETR has better detection results on the HOIST dataset itself.
Additional qualitative results, including failure cases, are provided in the appendix.

\noindent\textbf{Hand Detection Generalisation.} While HOI-DETR is primarily designed to predict complex hand-object interactions, we also evaluate its capability as a pure hand detector against specialised models in Table~\ref{tab:hand_detection_generalization}. Despite being optimised for the broader HOI task, HOI-DETR achieves the best hand detection performance on 4 out of the 6 evaluated datasets. Furthermore, HOI-DETR demonstrates superior zero-shot transfer on 3 out of 5 zero-shot datasets, achieving an average zero-shot AP of $82.4$ compared to WiLOR's $78.5$.

\begin{table}[t]
\centering
\setlength{\tabcolsep}{10pt}
\begin{tabular}{llll}
\toprule
Method & AP$_{50}$ $\uparrow$ & F1$_{\text{\textcolor{myred}{h}}\to\text{\textcolor{myyellow}{1st}}}$ ($\Delta$) $\uparrow$ & F1$_{\text{\textcolor{myyellow}{1st}}\to\text{\textcolor{mycyan}{2nd}}}$ ($\Delta$) $\uparrow$ \\
\midrule
No pretraining                  & 52.7 (\textcolor{red}{$-$33.4}) & 92.9 (\textcolor{red}{$-$2.7})  & 82.2 (\textcolor{red}{$-$12.2}) \\
No interaction module           & 86.0 (\textcolor{red}{$-$0.1})  & --                              & --                              \\
\quad + spatial heuristic (IoU) & 86.0 (\textcolor{red}{$-$0.1})  & 93.0 (\textcolor{red}{$-$2.6})  & 75.2 (\textcolor{red}{$-$19.2}) \\
Freeze backbone + encoder       & 71.6 (\textcolor{red}{$-$14.5}) & 93.6 (\textcolor{red}{$-$2.0})  & 93.1 (\textcolor{red}{$-$1.3})  \\
Freeze backbone                 & 82.7 (\textcolor{red}{$-$3.4})  & 95.5 (\textcolor{red}{$-$0.1})  & 94.0 (\textcolor{red}{$-$0.4})  \\
All pair interactions           & \textbf{86.1}                   & 94.7 (\textcolor{red}{$-$0.9})  & 93.3 (\textcolor{red}{$-$1.1})  \\
HOI-DETR (ours)        & \textbf{86.1}                   & \textbf{95.6}                   & \textbf{94.4}                   \\
\bottomrule
\end{tabular}

\vspace{1.0ex}
\caption{Ablation study on detection and interaction performance. We report the effect of pretraining, backbone freezing, and interaction modelling on detection performance and interaction performance (AP$_{50}$, F1$_{\text{\textcolor{myred}{h}}\to\text{\textcolor{myyellow}{1st}}}$, and F1$_{\text{\textcolor{myyellow}{1st}}\to\text{\textcolor{mycyan}{2nd}}}$). $\Delta$ denotes the absolute difference relative to HOI-DETR.}
\label{tab:ablation}
\end{table}

\begin{table}[h!]
\centering

\small
\begin{tabular*}{\columnwidth}{@{\extracolsep{\fill}}l c c l l@{}}
\toprule
Configuration & learned int. &  end-to-end &$\text{F1}_{\text{\textcolor{myred}{h}}\to\text{\textcolor{myyellow}{1st}}}$ ($\Delta$) $\uparrow$ & $\text{F1}_{\text{\textcolor{myyellow}{1st}}\to\text{\textcolor{mycyan}{2nd}}}$ ($\Delta$) $\uparrow$ \\
\midrule
Spatial heuristic (Nearest) & $\times$ &$\checkmark$ &  86.1 (\textcolor{red}{-9.5}) & 71.6 (\textcolor{red}{-22.8}) \\
Spatial heuristic (IOU)     & $\times$ &$\checkmark$ &  93.0 (\textcolor{red}{-2.6}) & 75.2 (\textcolor{red}{-19.2}) \\
Decoupled training     & $\checkmark$     & $\times$      & 81.2 (\textcolor{red}{-14.4}) &75.3 (\textcolor{red}{-19.1}) \\
Joint training (ours)       & $\checkmark$ & $\checkmark$ & \textbf{95.6}  & \textbf{94.4} \\
\bottomrule
\end{tabular*}
\vspace{-2ex}
\caption{\text{Interaction module ablation.} We evaluate the necessity of the learned 
interaction module and end-to-end training. $\Delta$ denotes absolute drop in F1 
relative to joint training. F1$_{\text{\textcolor{myred}{h}}\to\text{\textcolor{myyellow}{1st}}}$ and F1$_{\text{\textcolor{myyellow}{1st}}\to\text{\textcolor{mycyan}{2nd}}}$ 
report per-relation performance across all images; \textcolor{myyellow}{1st}$\to$\textcolor{mycyan}{2nd} pairs correspond to 
complex tool-use interactions requiring contextual chain reasoning.}
\label{tab:interaction_methods}
\end{table}

\noindent\textbf{Ablations.}
To understand the contribution of different components, we perform an ablation measuring both the detection performance (AP$_{50}$) and the interaction performance (F1$_{\text{\textcolor{myred}{h}}\to\text{\textcolor{myyellow}{1st}}}$, and F1$_{\text{\textcolor{myyellow}{1st}}\to\text{\textcolor{mycyan}{2nd}}}$)). Table~\ref{tab:ablation} reports results under varying configurations. Pretraining is absolutely critical to detection (-33.4 AP), as is tuning the transformer encoder that sits on top of the backbone (-14.5). Freezing the ViT backbone moderately hurts performance (-3.4). Training on all pairs of interactions (as opposed to selected pairs)  decreases the F1$_{\text{\textcolor{myred}{h}}\to\text{\textcolor{myyellow}{1st}}}$, and F1$_{\text{\textcolor{myyellow}{1st}}\to\text{\textcolor{mycyan}{2nd}}}$ by around 1 point. 

We also ablate the design and configuration of our interaction module in Table~\ref{tab:interaction_methods}. 
We compare our learned module with heuristic spatial rules (nearest box or max IoU). This substantially degrades performance, especially for complex \textcolor{myyellow}{\text{1st object}}$\to$\textcolor{mycyan}{\text{2nd object}} interactions where F1 drops (-$22.8\%$), confirming that spatial proximity is insufficient for contextual chain reasoning. We also compare end-to-end training with decoupled training. Decoupled training performs substantially worse across both relation types, dropping performance ($-14.4\%$ for \textcolor{myred}{\text{hand}}$\to$\textcolor{myyellow}{\text{1st object}} and -$19.1\%$ for \textcolor{myyellow}{\text{1st object}}$\to$\textcolor{mycyan}{\text{2nd object}}) compared to our joint training approach.

\section{Conclusion}
\label{sec:conclusion}

We have presented improvements to hand-object interaction on three fronts: a model (HOI-DETR), new evaluation data (a refined Hands23 and new HD-EPIC-HOI), and a comprehensive evaluation of HOI performance across images and video. 
Our end-to-end HOI-DETR model predicts interactions between all visible hands and any objects in a given image, whether directly or through a tool. 
Tested in a zero-shot setting on established benchmarks for hand and hand-object detection, results demonstrate robustness to challenging scenes (e.g. transparent lab equipment in FineBio) and complex interactions (e.g. hands in contact with multiple objects simultaneously).
Using the comprehensive evaluation set, we showed that the frame-based HOI-DETR strongly outperforms prior methods by over 20\% on both in- and out-of-distribution benchmarks. 

\paragraph{Acknowledgments}
This work was supported by EPSRC Program Grant Visual AI\hfill\break (EP/T028572/1). A Darkhalil was supported by EPSRC Doctoral Training Program (DTP). D. Fouhey was supported by the National Science Foundation under Grant No. 2006619 and 2437330.

We acknowledge the usage of GPU Node
hours granted as part of the  AIRR Innovator project ``5D Hand-Object Interaction Modelling from In-the-wild Videos'' (Mar 2026 - Sep 2026), AIRR Gateway project ``HOI Foundational Model from Egocentric Data'' (Dec 2025 - Mar 2026) and the Sovereign AI Unit call project
``Gen Model in Ego-sensed World'' (Aug 2025 - Nov 2025). 

We thank Sidhartha Reddy Potu for his contributions during the early stages of this project. We also gratefully acknowledge the authors of Co-DETR for open-sourcing their codebase and architecture, and the creators of the Hands23 dataset.

\bibliography{egbib}
\clearpage
\appendix
\noindent{\Large \textbf{Appendix}}
\vspace{1em}

We provide additional details and analyses.
\Cref{sec:supp_demo} introduces our public interactive demo and showcases HOI-DETR's robustness on challenging in-the-wild videos.
In \Cref{sec:supp_hands23}, we describe the refinement pipeline applied to the Hands23~\cite{cheng2023towards}
annotations, including the human verification setup, the automatic merging procedure for duplicated
boxes, and the resulting dataset statistics.
Similarly, \Cref{sec:hd_epic_hoi} details the manual quality-checking for masks of HD-EPIC-HOI, reporting frame- and sequence-level verification.
\Cref{sec:supp_quant} presents extended quantitative results: we analyse the effect of refinement
on Hands23, and report ablations on the interaction loss weight
$\lambda_{2}$, detector pretraining choices, and decoder-layer supervision.
Moroever, \Cref{sec:supp_qual} provides additional frame-level and video-level qualitative examples
on Hands23, HOIST, HD-EPIC-HOI, and web images.

\section{Interactive Demo and In-the-Wild Examples}
\label{sec:supp_demo}

To support adoption by the community, we release an interactive web demo of HOI-DETR alongside an extended set of qualitative results on challenging in-the-wild videos. Both are available from our \href{https://ahmaddarkhalil.github.io/HOI-DETR/}{project webpage}.

\begin{figure}[t]
    \centering
    \includegraphics[width=\linewidth]{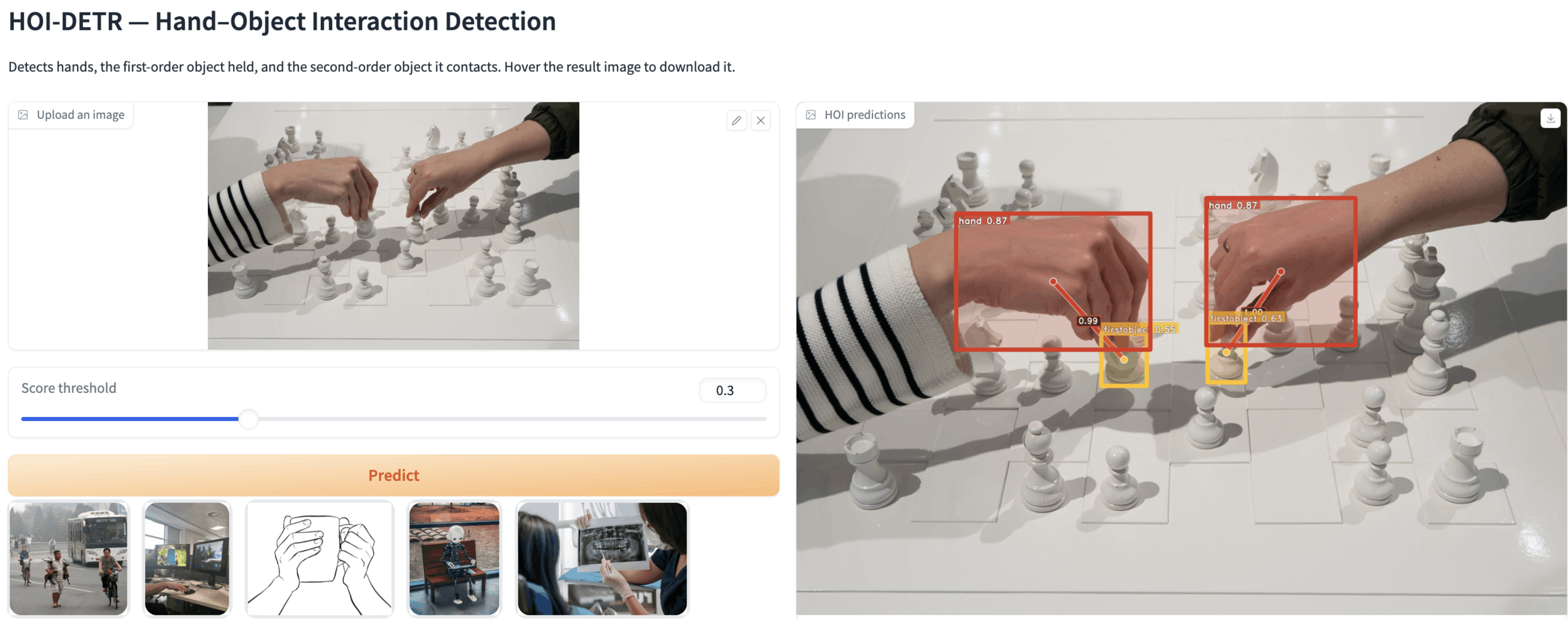}
    \caption{\textbf{Interactive Hugging Face demo.} Users upload an arbitrary image, adjust the score threshold, and inspect HOI-DETR predictions of \textcolor{myred}{hands}, \textcolor{myyellow}{1st objects}, and \textcolor{mycyan}{2nd objects} together with their interaction links. Example images are provided for quick exploration. Available at \href{https://huggingface.co/spaces/ahmaddarkhalil/hoi-detr-demo}{Hugging Face demo}.}
    \label{fig:demo_interface}
\end{figure}

\noindent\textbf{Interactive Hugging Face Demo.}
We host a \href{https://huggingface.co/spaces/ahmaddarkhalil/hoi-detr-demo}{Hugging Face demo} that allows users to upload an arbitrary image and evaluate HOI-DETR (Figure~\ref{fig:demo_interface}). The demo predicts \textcolor{myred}{hands}, \textcolor{myyellow}{1st objects}, and \textcolor{mycyan}{2nd objects} together with the interaction links between them, and the interface also has an adjustable score threshold so users can inspect predictions across a range of confidence levels. A panel of example images covering everyday scenes, line drawings, and out-of-distribution content is provided for quick exploration without requiring uploads.

\begin{figure}[t]
    \centering
    \includegraphics[width=\linewidth]{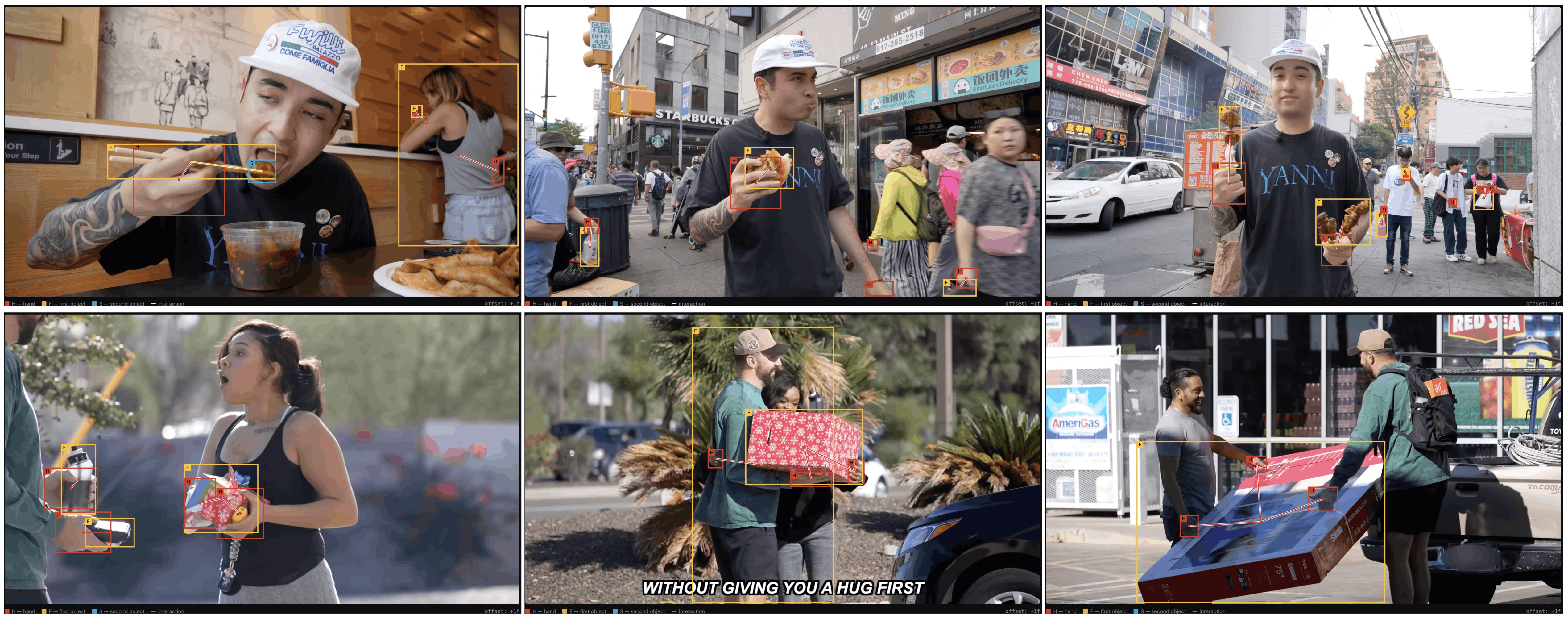}
    \caption{\textbf{HOI-DETR on challenging in-the-wild videos.} Each row corresponds to a different YouTube clip. HOI-DETR robustly detects \textcolor{myred}{hands}, \textcolor{myyellow}{1st objects}, and \textcolor{mycyan}{2nd objects} together with their interaction links across cluttered scenes, varied skin tones, and social interactions. Full video sequences are hosted on the \href{https://ahmaddarkhalil.github.io/HOI-DETR/}{project webpage}.}
    \label{fig:in_the_wild_videos}
\end{figure}

\noindent\textbf{In-the-Wild Videos.}
We further test HOI-DETR on a collection of YouTube videos with challenging hand-object interactions (Figure~\ref{fig:in_the_wild_videos}). These videos span street-food vending, gift exchanges, package handling, and casual social interactions that represent zero-shot cases, yet HOI-DETR produces consistent and accurate detections of \textcolor{myred}{hands}, \textcolor{myyellow}{1st objects}, and \textcolor{mycyan}{2nd objects} together with the corresponding interaction links throughout each sequence. The full set of annotated video sequences is available on the project webpage.

\section{Refined Hands23 Annotations}
\label{sec:supp_hands23}

This section provides a detailed description of the annotation refinement pipeline for the Hands23~\cite{cheng2023towards} dataset.
The goal of the refinement process is to resolve duplicated object annotations that arise from the Hands23 labeling protocol where each hand-object or object-object relation were annotated independently from others.
Additionally, we ensure consistent hand--object interaction links, and improve the overall quality of the dataset.
The following subsections describe the pipeline stages, the human verification setup, the automatic merging scheme for high-overlap candidates, 
and the resulting statistics after refinement.

\subsection{Refinement  Pipeline}
\label{sec:supp_hands23_pipeline}

The refinement pipeline consists of three main stages:
(i)~generation of images for refinement -- these contain pairs of bounding boxes with Intersection over Union (IoU) above a threshold and belonging to the same category,
(ii)~manual verification of candidate pairs within moderate IoU ranges, and
(iii) Automatic merging for high-overlap cases in the training set.
All stages preserve the internal consistency of interaction links by remapping interaction links 
whenever duplicate boxes are merged.

\begin{figure}[t]
    \centering
    \begin{minipage}[t]{0.54\linewidth}
        \centering
        \includegraphics[width=\linewidth]{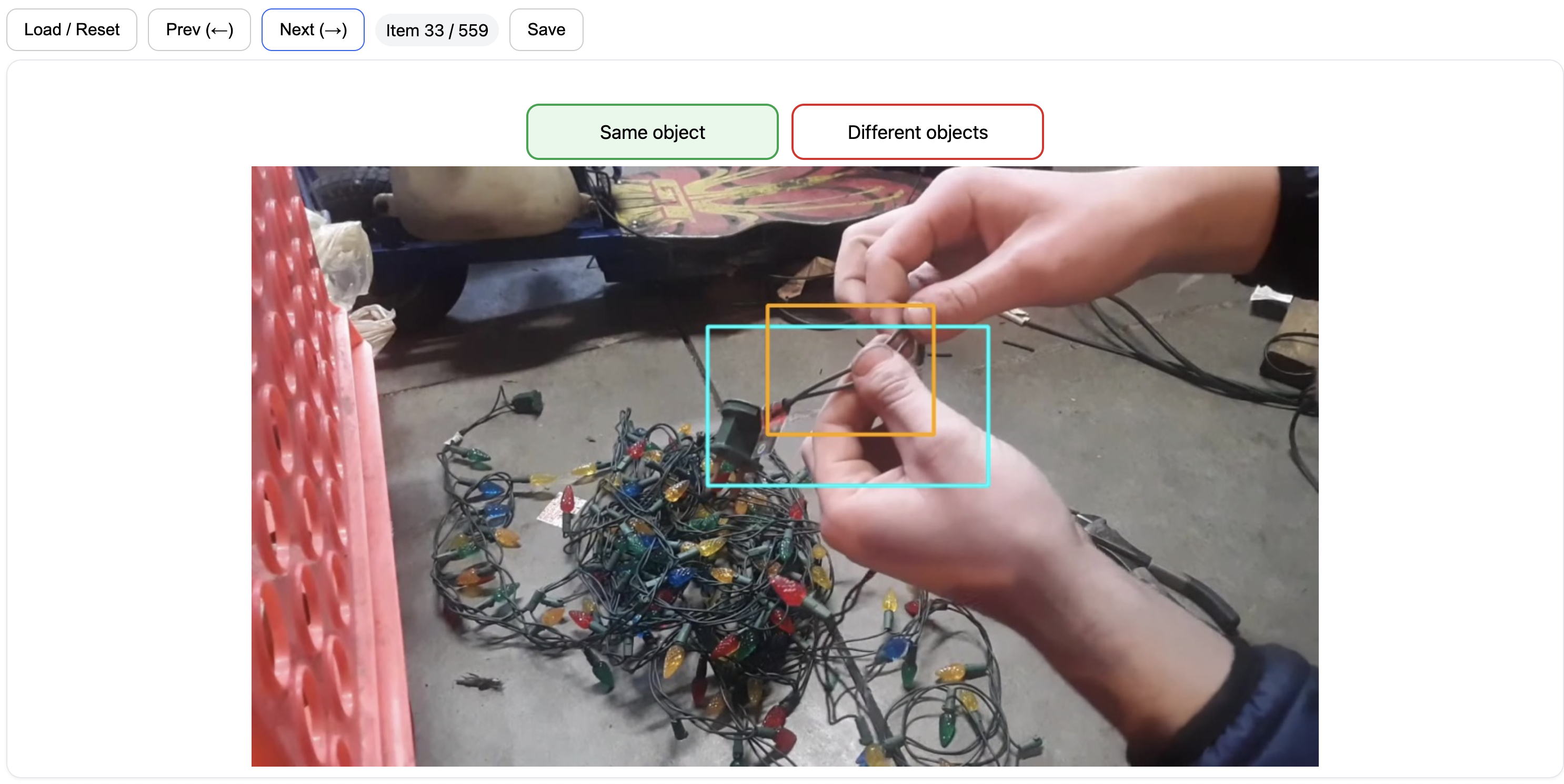}
        \caption{Annotation verification interface. Annotators are shown candidate box pairs (orange and cyan) and decide whether they correspond to the same physical object.}
        \label{fig:annotation_tool}
    \end{minipage}
    \hfill
    \begin{minipage}[t]{0.44\linewidth}
        \centering
        \includegraphics[width=\linewidth]{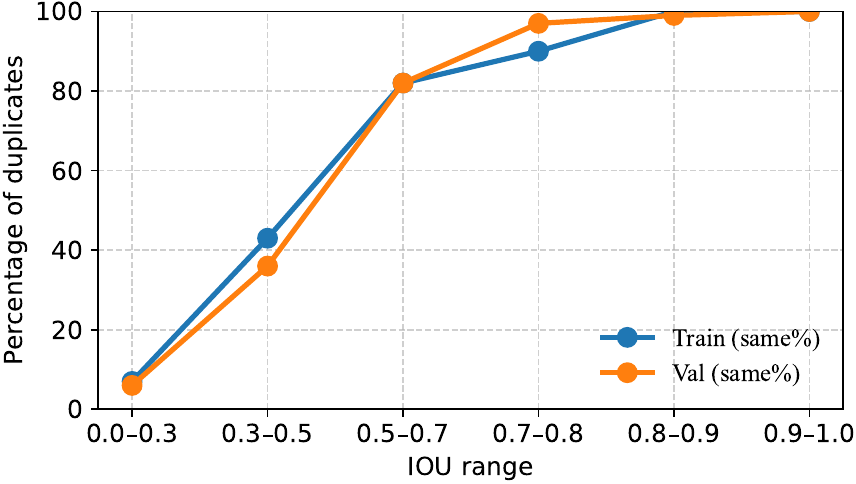}
        \caption{Percentage of duplicate pairs per IoU range. Duplicate frequency increases substantially in the 0.3--0.8 IoU interval.}
        \label{fig:iou_vs_merge}
    \end{minipage}
\end{figure}

\noindent\textbf{Human verification.}
To adjudicate ambiguous cases, we developed a web-based annotation interface (see Figure~\ref{fig:annotation_tool}).
For each image, annotators are shown a candidate pair of boxes and asked to determine whether they correspond to the 
\textbf{same physical object} or to \textbf{different instances}.
Boxes are visualised with distinct colours, and the tool provides keyboard shortcuts to support rapid and consistent annotation.
We partnered with a professional annotation company specialising in machine learning and computer-vision tasks.
A team of eight annotators carried out the verification, operating under written guidelines supplemented with examples of 
common edge cases (i.e. occlusion and partial visibility).
All verification was conducted in a controlled environment; no crowd-sourcing platforms were used.

\noindent\textbf{Manual and automatic merging.}
The entire validation split was manually verified 
    ($6{,}475$ candidate pairs).
For the training set, the refinement process separates the verification workload into two complementary components.

\begin{itemize}
    \vspace{-2pt}
    \item \textit{Manual verification (IoU 0.1--0.8).}  
    All candidate pairs in the training set within this IoU range were manually reviewed using the annotation tool.
    This covered $24{,}521$ candidate pairs.

    \item \textit{Automatic merging (IoU $>$ 0.8).} 
    For any pair with a very high overlap, we opted to merge these automatically as these were all incorrectly annotated from observing a random sample.
    A single instance bounding box is retained (chosen at random), and all interaction links are updated to point to the retained instance.
    \textbf{Importantly, no automatic merging is applied to the validation set}.
\end{itemize}

\noindent\textbf{Correction frequency by IoU.}
Figure~\ref{fig:iou_vs_merge} shows the percentage of candidate pairs annotated as duplicates within each IoU bin.
Both training and validation sets exhibit a sharp increase in duplicate frequency between 0.3--0.8 IoU, 
highlighting the ambiguity of moderate-overlap regions.
High IoU pairs ($>$0.8) show near-perfect duplicate agreement in the validation set, 
justifying the use of automatic merging for the training data in this regime.

\subsection{Refined Hands23 Statistics}
\label{sec:supp_hands23_stats}
We next quantify the effect of our refinement procedure on the overall dataset scale.
Table~\ref{tab:supp_hands23_stats} summarises the refined annotation statistics 
for Hands23 across its four constituent subsets, together with the absolute change 
relative to the original Hands23.
The table provides a clear reference for how the dataset is affected by our 
correction pipeline.
As expected, the most substantial reductions occur in the~\textcolor{myyellow}{1st objects} category 
(37.2k in train and 4.4k in val), reflecting that most duplicate cases arise when 
a single object is jointly manipulated by two hands, leading to two overlapping 
annotations for the same physical item.
Hand annotations remain effectively unchanged, as they were already reliable and 
not subject to our refinement.

\begin{table}[t]
\centering
\setlength{\tabcolsep}{2pt}
\begin{adjustbox}{width=\linewidth}
\begin{tabular}{lcccccccc}
\toprule
\multirow{2}{*}{Subset}
& \multicolumn{4}{c}{Train (refined, $\times 10^3$)}
& \multicolumn{4}{c}{Val (refined, $\times 10^3$)} \\
\cmidrule(lr){2-5} \cmidrule(lr){6-9}
& \#Images & \#Hands & \#1st obj & \#2nd obj
& \#Images & \#Hands & \#1st obj & \#2nd obj \\
\midrule
EK VISOR (EK)~\cite{darkhalil2022epic}
& 25.2 & 38.3
& 28.6 {\scriptsize\textcolor{blue!60!black}{($-3.4$)}}
& 4.0  {\scriptsize\textcolor{blue!60!black}{($-0.2$)}}
& 6.3  & 9.5
& 7.1  +{\scriptsize\textcolor{blue!60!black}{($-0.9$)}}
& 1.0  {\scriptsize\textcolor{blue!60!black}{($-0.1$)}} \\

New Days (ND)~\cite{cheng2023towards}
& 75.8 & 95.6
& 60.0 {\scriptsize\textcolor{blue!60!black}{($-13.5$)}}
& 7.2  {\scriptsize\textcolor{blue!60!black}{($-0.5$)}}
& 10.9 & 14.3
& 9.2  {\scriptsize\textcolor{blue!60!black}{($-1.9$)}}
& 1.3  {\scriptsize\textcolor{blue!60!black}{($-0.1$)}} \\

Articulation (AR)\cite{Qian22}
& 68.7 & 88.3
& 49.9 {\scriptsize\textcolor{blue!60!black}{($-9.1$)}}
& 0.7  {\scriptsize\textcolor{blue!60!black}{($-0.1$)}}
& 2.8  & 3.5
& 2.2  {\scriptsize\textcolor{blue!60!black}{($-0.3$)}}
& 0.04  \\

COCO (CC)~\cite{lin2014microsoft}
& 36.6 & 
99.2
& 52.8 {\scriptsize\textcolor{blue!60!black}{($-11.1$)}}
& 1.6  {\scriptsize\textcolor{blue!60!black}{($-0.2$)}}
& 4.6  & 12.4
& 6.6  {\scriptsize\textcolor{blue!60!black}{($-1.3$)}}
& 0.2  \\

\midrule
Total
& 206.3 & 321.4
& 191.4 {\scriptsize\textcolor{blue!60!black}{($-37.2$)}}
& 13.6 {\scriptsize\textcolor{blue!60!black}{($-0.9$)}}
& 24.6 & 39.8
& 25.0 {\scriptsize\textcolor{blue!60!black}{($-4.4$)}}
& 2.5  {\scriptsize\textcolor{blue!60!black}{($-0.2$)}} \\
\bottomrule
\end{tabular}
\end{adjustbox}
\vspace{1.5ex}
\caption{Statistics of Hands23 before and after refinement.
We report the number of images and instances per subset for the refined
annotations. Values are expressed in units of $10^3$ instances.
Numbers in brackets indicate the absolute change with respect to the
original annotations (refined $-$ original). Negative values correspond
to removed instances due to redundancy.}
\label{tab:supp_hands23_stats}
\end{table}

\FloatBarrier
\section{HD-EPIC-HOI Annotation Quality}
\label{sec:hd_epic_hoi}
To ensure the HD-EPIC-HOI annotations are of high quality, we perform two manual checks: a verification of the sequences and an independent assessment of the SAM2~\cite{ravi2025sam} mask propagations.

\begin{figure}[t]
    \centering
    \begin{minipage}[t]{0.485\linewidth}
        \centering
        \includegraphics[width=\linewidth]{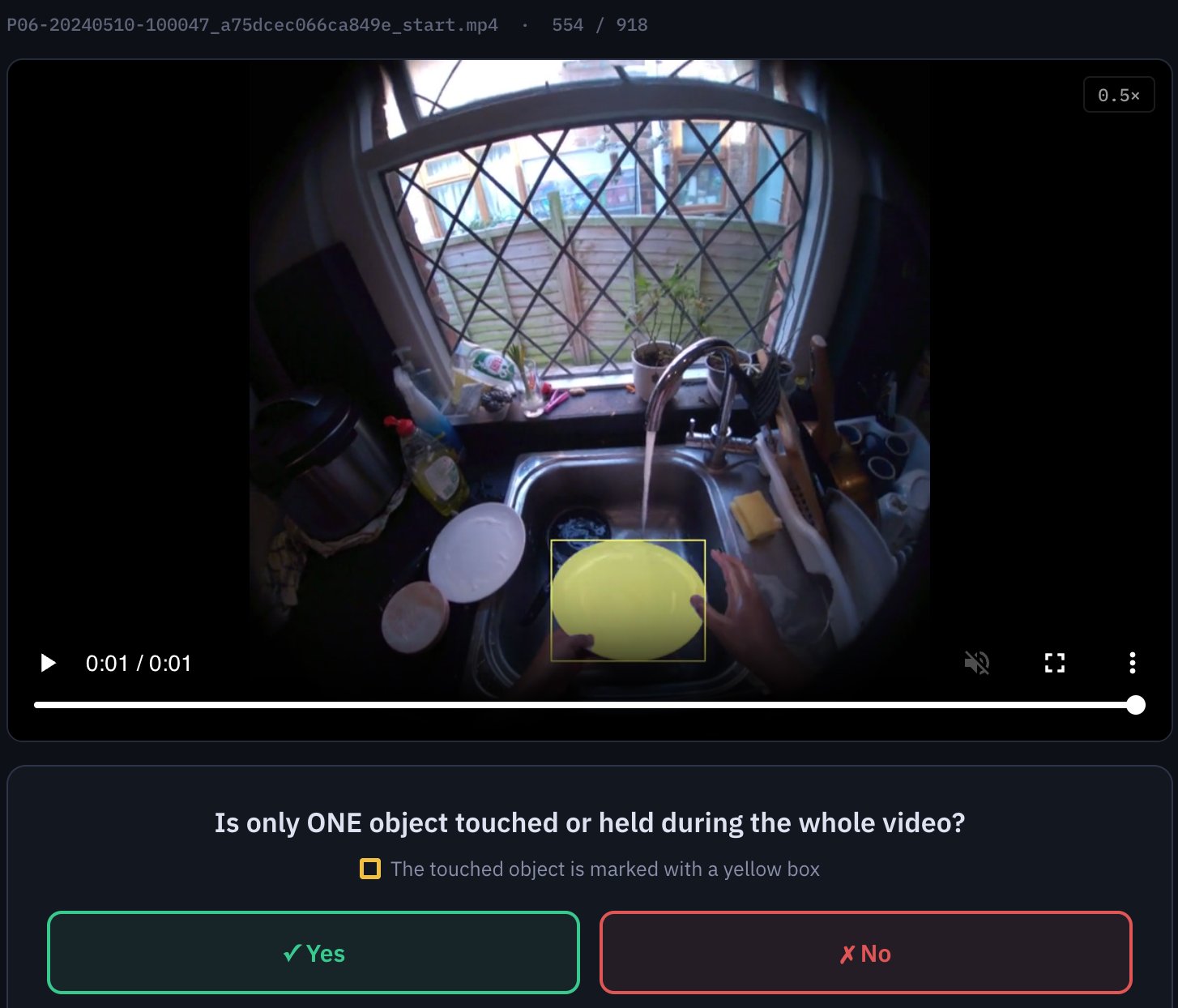}
        \caption{Single-object verification interface. For each sequence, the annotator watches the video and decides whether the highlighted object (yellow box) is the only object touched throughout the clip.}
        \label{fig:single_obj_interface}
    \end{minipage}
    \hfill
    \begin{minipage}[t]{0.50\linewidth}
        \centering
        \includegraphics[width=\linewidth]{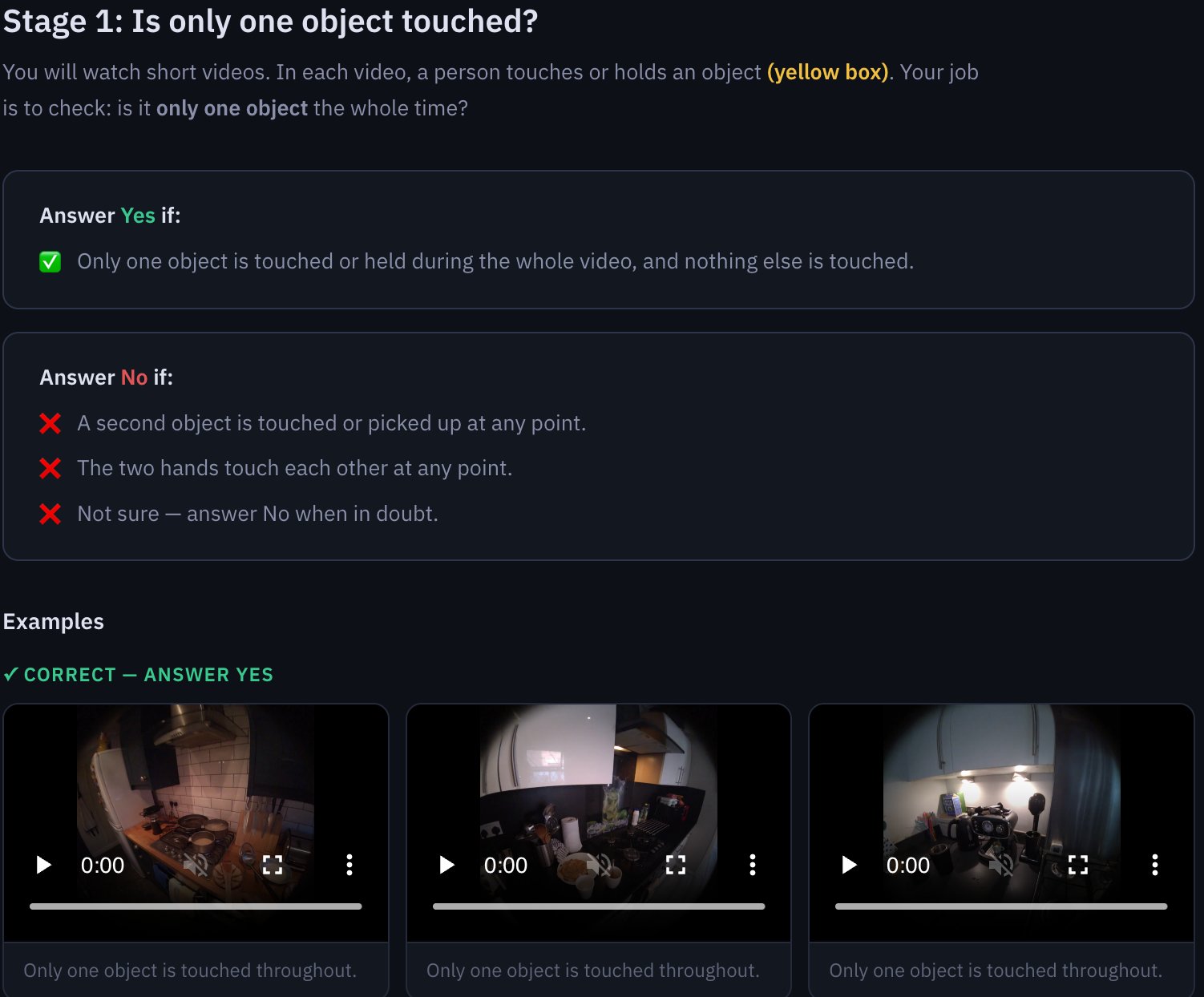}
        \caption{Instructions provided to annotators. A sequence is accepted only when a single annotated object is touched or held in the entire video; any contact with another object or hand-to-hand contact triggers rejection.}
        \label{fig:single_obj_instructions}
    \end{minipage}
\end{figure}

\noindent\textbf{Manual Verification of HD-EPIC-HOI Sequences}.
To ensure the quality of HD-EPIC-HOI annotations, we manually verify that each sampled sequence includes interaction with only a single annotated object. For each sequence, an annotator watches the clip and decides whether the annotated object is the only object the hands interact with, i.e.\ no other object is touched, picked up, or manipulated throughout the clip. The verification interface and the instructions provided to annotators are shown in Figures~\ref{fig:single_obj_interface} and~\ref{fig:single_obj_instructions}. Out of 911 initially sampled sequences, this filtering retains 768 (35.3k images and 22.2k \textcolor{myyellow}{1st object} bounding boxes), removing noise from hand contact with additional objects and yielding clean single-object HOI suitable for evaluating \textcolor{myyellow}{1st object} detection.

\begin{figure}[t]
    \centering
    \begin{minipage}[t]{0.44\linewidth}
        \centering
        \includegraphics[width=\linewidth]{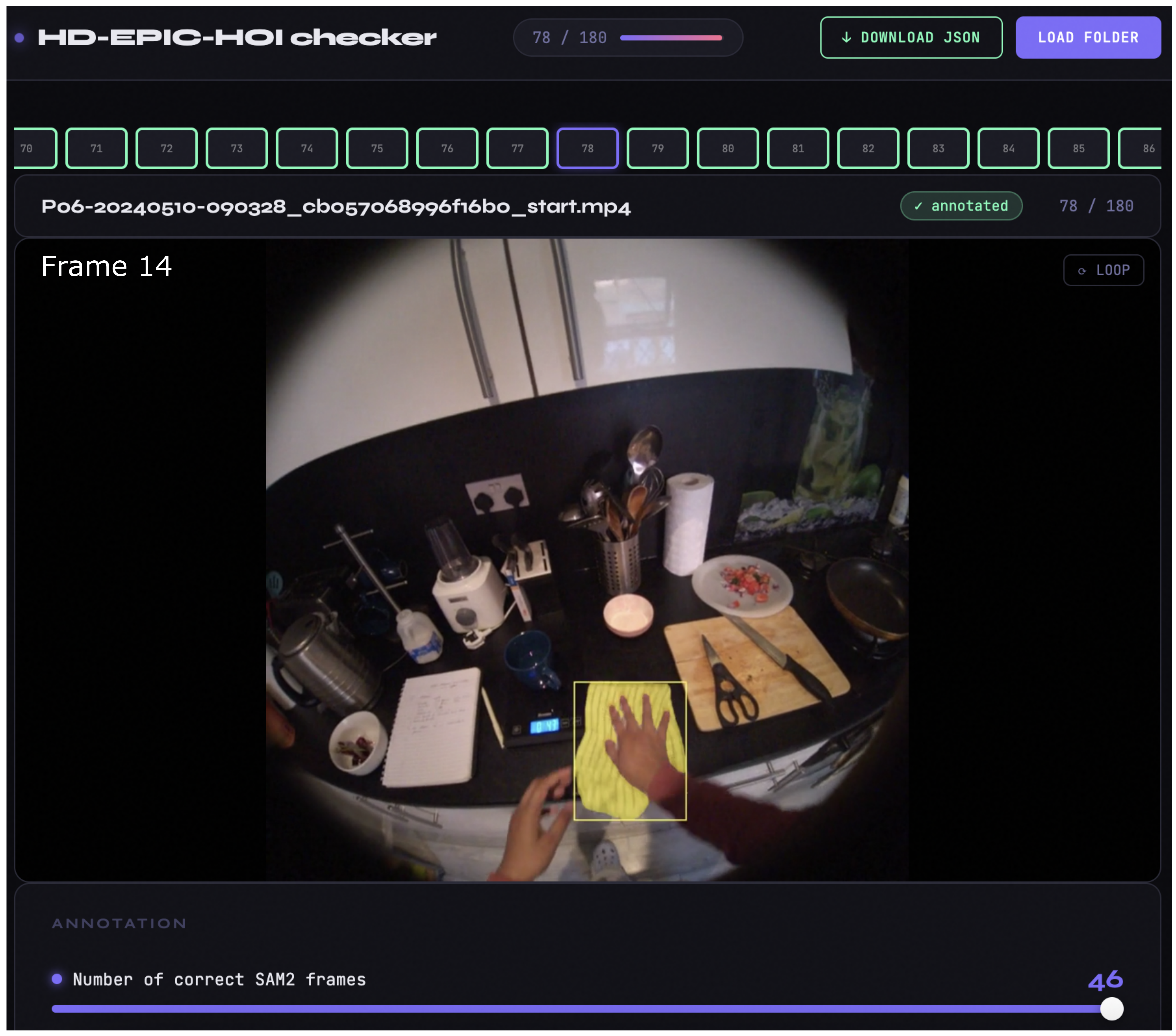}
        \caption{SAM2 quality-checking interface. For each sequence, the annotator watches the looping video and records the number of frames where SAM2 correctly propagated the object mask.}
        \label{fig:hd_epic_interface}
    \end{minipage}
    \hfill
    \begin{minipage}[t]{0.53\linewidth}
        \centering
        \includegraphics[width=\linewidth]{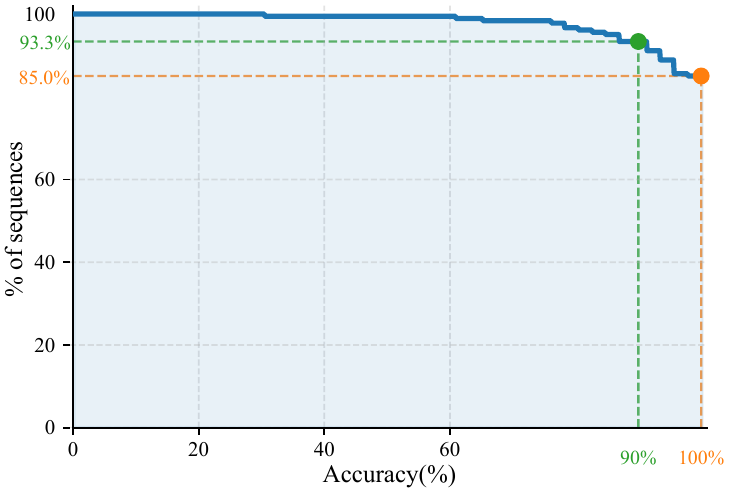}
        \caption{SAM2 propagation quality across sequences. The curve shows the percentage of sequences achieving at least a given sequence-level accuracy. 85\% of sequences are propagated perfectly (100\%), and 93.3\% achieve at least 90\% accuracy.}
        \label{fig:hd_epic_stats}
    \end{minipage}
\end{figure}

\noindent\textbf{SAM2 Annotation Quality}.
We also assess the per-frame quality of the SAM2~\cite{ravi2025sam} propagations through an independent manual check on a random sample of 180 sequences ($\sim$23\% of the dataset), covering 8{,}280 frames in total. For each sequence, an annotator watches the full video multiple times and records the number of frames where SAM2 correctly propagated the object mask, using the interface shown in Figure~\ref{fig:hd_epic_interface}.

Overall, \textbf{97.83\%} of frames are correctly propagated (8{,}100 out of 8{,}280). At the sequence level defined as the fraction of correctly propagated frames within a sequence, \textbf{85\%} of sequences are propagated perfectly across all frames and \textbf{93.3\%} achieve at least 90\% per-sequence accuracy (Figure~\ref{fig:hd_epic_stats}). We attribute this robustness to two design choices: (i) using accurate HD-EPIC start/end timestamps that tightly bound the period of object motion, and (ii) applying SAM2 only over short segments, limiting temporal inconsistency or drift. Taken together with the single-object verification, these results confirm the high quality of the HD-EPIC-HOI annotations, which are released as-is without any further correction.

\section{Additional Quantitative Results}
\label{sec:supp_quant}

We provide extended quantitative results that complement the main paper:
more detailed Hands23 analyses, per-dataset breakdowns and ablations.

\subsection{Extended Hands23 evaluation}
\label{sec:supp_hands23_extended}
The results in Table~\ref{tab:supp_refined_hands23_scales} show that HOI-DETR consistently and substantially outperforms the Hands23 detector across all object scales when evaluated on the refined Hands23 annotations.
The key observation is that the improvement is almost uniform across scales: for small objects, AP$_{50}$ increases from 26.4 to 47.6 (+21.2);
for medium objects, AP$_{50}$ improves from 52.0 to 73.7 (+21.7);
and for large objects, it rises from 70.0 to 90.1 (+20.1).
Although absolute performance remains highest for large objects, this near-constant gain means that the most challenging small instances, such as hand-held tools and utensils that occupy only a small fraction of the image, benefit proportionally the most from the refined annotations and the HOI-DETR architecture.

As Hands23 was made up by accumulating images from multiple dataset, we report performance for each of the constituting datasets, which we refer to as subsets.
Table~\ref{tab:supp_refined_hands23_subsets} further breaks down performance by Hands23 subset (EK, ND, CC, AR), again evaluated on the refined annotations.
HOI-DETR improves over the Hands23 detector on every subset, but the magnitude of the gain varies with the underlying domain.
On the EK subset (egocentric), AP$_{50}$ rises from 78.2 to 91.6 (+13.4), yielding already high absolute performance on a domain where the baseline is strong.
The improvements are even more pronounced on ND and AR, where HOI-DETR reaches 87.0 and 87.1 AP$_{50}$, corresponding to gains of 24.5 and 26.3 points, respectively.
The largest gain is observed on the CC subset (COCO images), where AP$_{50}$ increases from 42.8 to 71.2 (+28.4), indicating substantially better generalisation to diverse third-person images with a wide range of object categories and viewpoints.

Taken together, these results suggest that the combination of refined Hands23 annotations and the HOI-DETR architecture yields a detector that is not only stronger overall but also more robust across domains and object scales than the original Hands23 model.

\begin{table}[t]
\centering
\begin{tabular}{lccc}
\toprule
Method                       & AP$_{50}^{\text{small}}$ & AP$_{50}^{\text{medium}}$ & AP$_{50}^{\text{large}}$ \\
\midrule
Hands23
& 26.4 & 52.0 & 70.0 \\
HOI-DETR
& \textbf{47.6} {\scriptsize\textcolor{green!50!black}{(+21.2)}}
& \textbf{73.7}{\scriptsize\textcolor{green!50!black}{(+21.7)}}
& \textbf{90.1}{\scriptsize\textcolor{green!50!black}{(+20.1)}} \\
\bottomrule
\end{tabular}
\vspace{1.5ex}
\caption{\textbf{Refined Hands23 evaluation.}
AP$_{50}$ on the refined Hands23 evaluation data for small, medium, and large objects, comparing HOI-DETR (trained on refined Hands23) and Hands23 (trained on the original Hands23).}
\label{tab:supp_refined_hands23_scales}
\end{table}

\begin{table}[t]
\centering
\setlength{\tabcolsep}{3pt} 
\begin{tabular}{lllll}
\toprule
Method    & EK & ND & AR & CC \\
\midrule
Hands23
& 78.2 & 62.5 & 60.8 &42.8  \\
HOI-DETR (Ours)
& \textbf{91.6} {\scriptsize\textcolor{green!50!black}{(+13.4)}}%
& \textbf{87.0} {\scriptsize\textcolor{green!50!black}{(+24.5)}}%
& \textbf{87.1} {\scriptsize\textcolor{green!50!black}{(+26.3)}} 
& \textbf{71.2 }{\scriptsize\textcolor{green!50!black}{(+28.4)}}%
\\
\bottomrule
\end{tabular}
\vspace{1.5ex}
\caption{\textbf{Refined Hands23 subset-level evaluation.}
Subset-wise AP$_{50}$ on the refined Hands23 evaluation data for each Hands23 subset.}
\label{tab:supp_refined_hands23_subsets}
\end{table}

\subsection{Ablation on Interaction Loss Weight \texorpdfstring{$\lambda_{2}$}{lambda2}}
\label{sec:supp_lambda2}

\begin{wraptable}{r}{0.32\textwidth}
    \centering
    \vspace{-1.5em}
    \setlength{\tabcolsep}{6pt}
    \small
    \begin{tabular}{lcc}
        \toprule
        $\lambda_{2}$ & AP$_{50}$ & $\mathrm{F1}_{\mathrm{inter}}$ \\
        \midrule
        0  & 86.0 & - \\
        1  & 86.1 & 94.9 \\
        3  & 86.1 & 95.5 \\
        6  & 85.9 & 95.7 \\
        9  & 85.8 & 95.8 \\
        12 & 85.6 & 95.8 \\
        15 & 85.7 & 95.8 \\
        18 & 85.4 & 95.8 \\
        \bottomrule
    \end{tabular}
    \vspace{1.5ex}
    \caption{Ablation of interaction loss weight $\lambda_{2}$.}
    \label{tab:supp_lambda2}
    \vspace{-1.0em}
\end{wraptable}
We study the effect of the interaction loss weight $\lambda_{2}$ in the global objective introduced in Eq.~(4) of the main paper, where $\mathcal{L}^{int}_{l}$ supervises the interaction predictions at each decoder layer.
As shown in Table~\ref{tab:supp_lambda2}, setting $\lambda_{2}=0$ (i.e., removing interaction supervision) yields an AP$_{50}$ of 86.0, and we omit the $\text{F1}_{\text{inter}}$ as the interaction head is not trained in this setting.
Once interaction supervision is enabled ($\lambda_{2}\geq 1$), the $\text{F1}_{\text{inter}}$ increases sharply, with most of the gains already realised by $\lambda_{2}=3$ and only marginal improvements beyond $\lambda_{2}=6$.
In contrast, very large values of $\lambda_{2}$ slightly bias the optimisation towards the interaction head and lead to a small but consistent decrease in AP$_{50}$.

Overall, these results indicate that the model is robust to the precise choice of $\lambda_{2}$ within a broad intermediate range: interaction performance quickly enters a plateau, while detection performance remains essentially unchanged as long as $\lambda_{2}$ is not set excessively high.
We select $\lambda_{2}=3$ as our default configuration, as it lies at the beginning of the plateau where $\text{F1}_{\text{inter}}$ is already close to its maximum, while maintaining the highest AP$_{50}$ values and avoiding the mild degradation observed for larger $\lambda_{2}$.

\subsection{Ablation on Pretraining}

Table~\ref{tab:supp_pretraining} reports an ablation on detector pretraining, where HOI-DETR is trained on the refined Hands23 training set and evaluated on the corresponding refined validation set.
Without pretraining, AP$_{50}$ drops dramatically to 52.7 with a corresponding decrease in $\text{F1}_{\text{inter}}$, whereas all three pretraining datasets (Objects365, LVIS, COCO) bring the model into a high-performance regime.
Among them, the differences are very small: Objects365 and COCO achieve identical AP$_{50}$ and $\text{F1}_{\text{inter}}$, and LVIS slightly improves $\text{F1}_{\text{inter}}$ at the cost of a small AP$_{50}$ decrease.
We therefore adopt COCO pretraining as our default configuration relying on a standard, widely used dataset that facilitates reproducibility and aligns well with the COCO-derived component of Hands23.

\begin{table}[t]
    \centering
    \begin{minipage}[t]{0.42\linewidth}
        \centering
        \begin{tabular}[t]{lcc}
        \toprule
        Pretraining & AP$_{50}$ & $\text{F1}_{\text{inter}}$ \\
        \midrule
        None & 52.7 & 91.5 \\
        LVIS~\cite{gupta2019lvis} & 85.7 & 95.7 \\
        Objects365~\cite{shao2019objects365} & 86.1 & 95.5 \\
        COCO~\cite{lin2014microsoft} & 86.1 & 95.5 \\
        \bottomrule
        \end{tabular}
        \vspace{1.5ex}
        \caption{Ablation of pretraining datasets.
        AP$_{50}$ and  $\text{F1}_{\text{inter}}$ for HOI-DETR with different detector pretraining configurations.}
        \label{tab:supp_pretraining}
    \end{minipage}
    \hfill
    \begin{minipage}[t]{0.55\linewidth}
        \centering
        \begin{tabular}[t]{lccc}
        \toprule
        Setting & $\lambda_{2}$ & AP$_{50}$ & $\text{F1}_{\text{inter}}$ \\
        \midrule
        Int.\ last layer only & 3  & 86.1 & 94.8 \\
        Int.\ last layer only & 18 & 85.7 & 95.3 \\
        Dec.\ losses last layer only & 3  & 85.8 & 95.2 \\
        All layers (ours) & 3  & 86.1 & 95.5 \\
        \bottomrule
        \end{tabular}
        \vspace{-1.5ex}
        \caption{Ablation on decoder-layer supervision.
        Impact of applying the interaction loss and decoder losses on all layers vs.\ only the final decoder layer.
        Int. refers to interaction loss.}
        \label{tab:supp_decoder_layers}
    \end{minipage}
\end{table}

\subsection{Architecture Ablation}
Table~\ref{tab:supp_decoder_layers} analyses how supervising different decoder layers affects performance.
Applying the interaction loss on all decoder layers (our default) yields the highest interaction $\text{F1}_{\text{inter}}$~(95.5) without sacrificing AP$_{50}$ compared to restricting interaction supervision to the final layer only.
Increasing $\lambda_{2}$ to 18 in the last-layer-only recovers most of $\text{F1}_{\text{inter}}$ but slightly degrades AP$_{50}$, and concentrating all decoder losses on the last layer also leads to a small drop in both metrics.
These results support the design choice of distributing detection and interaction supervision across decoder layers rather than focusing it solely on the final layer.

\section{Additional Qualitative Results on Benchmarks}
\label{sec:supp_qual}

\begin{figure}[t]
    \centering
    \includegraphics[width=\linewidth]{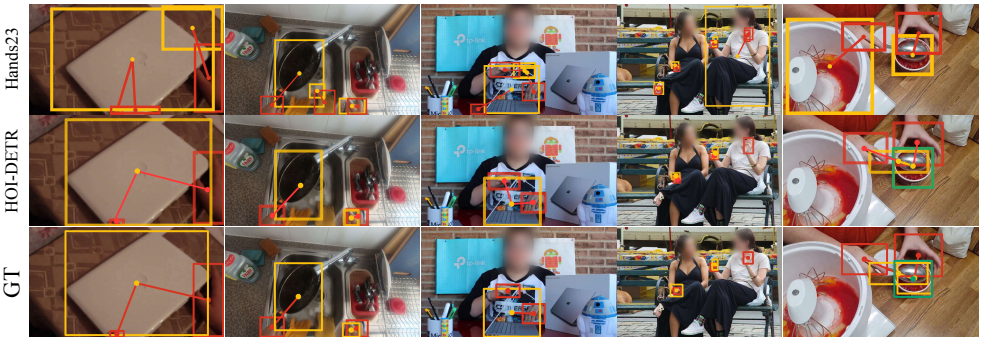}
    \caption{Qualitative results of HOI-DETR and Hands23 model on Hands23 dataset}
    \label{fig:qualititve_image_level}
\end{figure}

\begin{figure}[t]
    \centering
    \includegraphics[width=\linewidth]{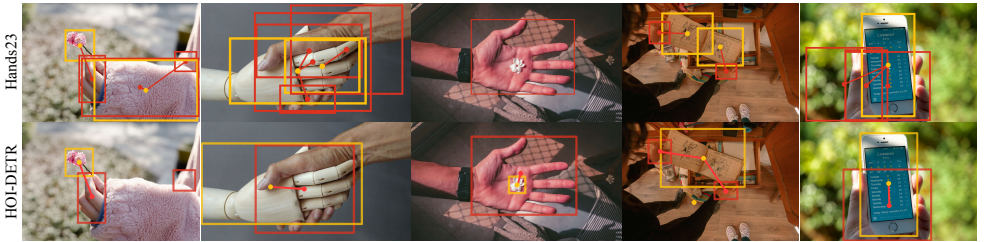}
    \caption{Qualitative results of HOI-DETR and Hands23 model on web images.}
    \label{fig:qualititve_image_level_zero}
\end{figure}

\begin{figure}[t]
    \centering
    \includegraphics[width=\linewidth]{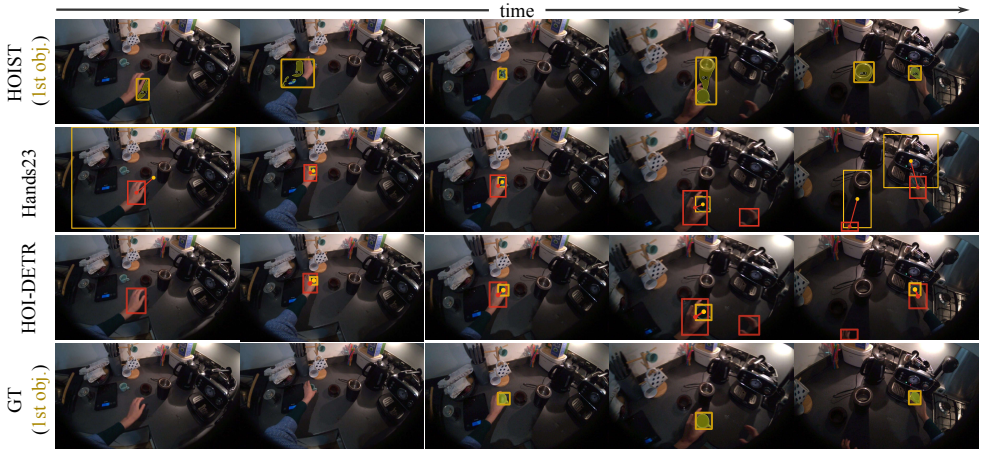}
    \caption{Qualitative video-sequence from HD-EPIC-HOI. We compare Hands23, HOIST and HOI-DETR.
    The ground truth (GT) annotations provide only the \textcolor{myyellow}{1st object} annotations, but we also show all predicted hand regions for both Hands23 and HOI-DETR.
    Boxes for HOIST are displayed on top of the mask for visualisation purposes.}
    \label{fig:qualitive_video_hd_epic_supp}
\end{figure}

We complement the quantitative analysis with additional frame-level and video-level qualitative examples on Hands23, HOIST, HD-EPIC-HOI, and web images.
Figure~\ref{fig:qualititve_image_level} shows image-level comparisons on Hands23, Figure~\ref{fig:qualititve_image_level_zero} illustrates zero-shot behaviour on diverse web images, and Figure~\ref{fig:qualitive_video_hd_epic_supp} presents a video sequence from HD-EPIC-HOI comparing Hands23, HOIST, and HOI-DETR.
In addition, we also show failure cases.

\begin{figure}[t]
    \centering
    \includegraphics[width=\linewidth]{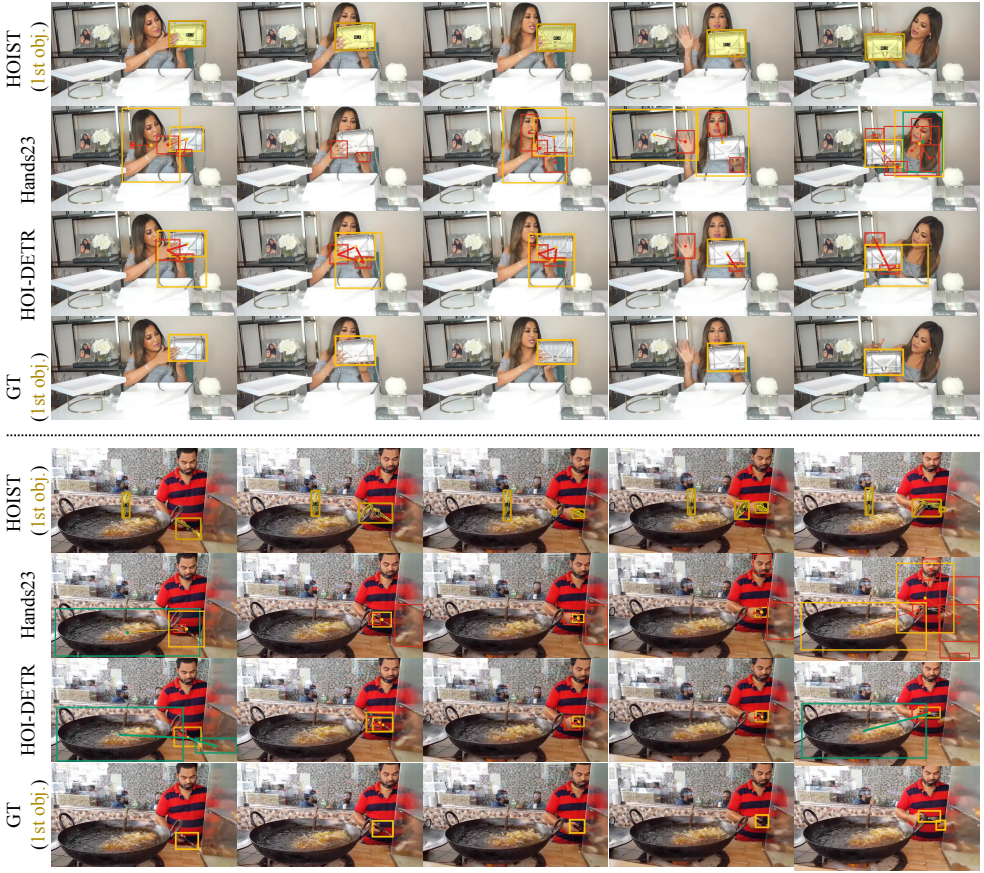}
    \caption{
        \textbf{Examples of video-level failure cases from HOIST.}
        We show two randomly selected video clips from the HOIST dataset and compare Hands23, HOIST and HOI-DETR against the ground-truth (GT) \textcolor{myyellow}{1st object} annotations.
        We also visualise all predicted hand regions for both Hands23 and HOI-DETR, while HOIST predictions are displayed as boxes overlaid on the masks for visualisation purposes.
    }
    \label{fig:qualitive_video_random_hoist_supp}
\end{figure}

\noindent\textbf{Failure Cases.}
Figure~\ref{fig:qualitive_video_random_hoist_supp} shows failure cases on HOIST.
In the first sequence, duplicate boxes capture both a ‘hand bag’ and its chain, treating them as separate detections.
In the second video sequence, the interaction links for the \textcolor{mycyan}{2nd object} are incorrect.
These samples illustrate remaining limitations of HOI-DETR, particularly around small or elongated objects, ambiguous object extent, and fine-grained associations, where close proximity between objects can still trigger false interaction detections despite the absence of physical contact.

\end{document}